\documentclass{article}

\usepackage{arxiv}

\usepackage[utf8]{inputenc} 
\usepackage[T1]{fontenc}    
\usepackage[arabic,english]{babel}
\usepackage{hyperref}       
\usepackage{cleveref}
\usepackage{url}            
\usepackage{booktabs}       
\usepackage{amsfonts}       
\usepackage{nicefrac}       
\usepackage{microtype}      
\usepackage{lipsum}		
\usepackage{graphicx}
\usepackage{natbib}
\usepackage{doi}
\usepackage[table]{xcolor}
\usepackage{makecell,multirow}

\usepackage{devanagari}

\usepackage{CJKutf8}

\usepackage{caption}
\usepackage{graphicx}
\graphicspath{ {./figures/} }

\usepackage{authblk}

\usepackage{amsthm}
\newtheorem{definition}{Definition}

\newcolumntype{C}[1]{>{\centering\let\newline\\\arraybackslash\hspace{0pt}}m{#1}}

\title{Crowdsourcing Lexical Diversity}

\author[1,2]{Hadi Khalilia}
\author[3]{Jahna Otterbacher}
\author[4]{G\'abor Bella}
\author[1]{Shandy Darma}
\author[1]{Fausto Giunchiglia}
\affil[1]{Department of Information Engineering and Computer Science, University of Trento, Trento, Italy}
\affil[2]{Palestine Technical University--Kadoorie, Palestine}
\affil[3]{Open University of Cyprus, CYENS - Centre of Excellence, Nicosia, Cyprus}
\affil[4]{Lab-STICC CNRS UMR 628, IMT Atlantique, Brest, France}

\date{}

\renewcommand{\headeright}{}
\renewcommand{\undertitle}{}
\renewcommand{\shorttitle}{Crowdsourcing Lexical Diversity}

\hypersetup{
pdftitle={Crowdsourcing Lexical Diversity},
pdfauthor={Hadi Khalilia, Jahna Otterbacher, G\'abor Bella, Shandy Darma, Fausto Giunchiglia},
pdfkeywords={multilingual lexicon, language diversity, crowdsourcing, linguistic gap, lexical typology},
}

\begin{document}
\maketitle

\begin{abstract}
Lexical-semantic resources (LSRs), such as online lexicons and wordnets, are fundamental to natural language processing applications as well as to fields such as linguistic anthropology and language preservation.
In many languages, however, such resources suffer from quality issues: incorrect entries, incompleteness, but also the rarely addressed issue of bias towards the English language and Anglo-Saxon culture. Such bias manifests itself in the absence of concepts specific to the language or culture at hand, the presence of foreign (Anglo-Saxon) concepts, as well as in the lack of an explicit indication of untranslatability, also known as cross-lingual \emph{lexical gaps}, when a term has no equivalent in another language. This paper proposes a novel crowdsourcing methodology for reducing bias in LSRs. Crowd workers compare lexemes from two languages, focusing on domains rich in lexical diversity, such as kinship or food. Our LingoGap crowdsourcing platform facilitates comparisons through microtasks identifying equivalent terms, language-specific terms, and lexical gaps across languages.  We validated our method by applying it to two case studies focused on food-related terminology: (1)~English and Arabic, and (2)~Standard Indonesian and Banjarese. These experiments identified 2,140~lexical gaps in the first case study and 951~ in the second. The success of these experiments confirmed the usability of our method and tool for future large-scale lexicon enrichment tasks.
\end{abstract}


\section{Introduction}


Despite advances in deep learning and large language models, lexical-semantic resources (LSRs)—such as WordNet \citep{miller1995}—remain essential for natural language processing (NLP) tasks, including machine translation, word sense disambiguation, and information retrieval \citep{katsuta2020l, loureiro2019, barbouch2021}, as well as for supporting broader research domains such as linguistic anthropology, cultural linguistics, and language documentation.



%

Due to its dominance, the English language and, in particular, Princeton WordNet (PWN) \citep{miller1995}, has played a distinguished role in the construction of lexical databases for many languages. The English lexicon has been widely adopted as a \textit{pivot} representation of lexical meaning across languages, but also as the source language for \textit{translation-based} lexicon development \citep{bond2013}. Relying on PWN as a ``standard" meaning inventory, however, leads to the creation of resources that suffer from \emph{language modeling bias}, due to deep-running linguistic and cultural differences across speaker communities \citep{giunchiglia2017understanding,giunchiglia2023representing,bella2024tackling}. For instance, English and Italian lack an equivalent equivalent for the Arabic word \AR{خالة}, which means ``\textit{mother's sister}'', whereas Arabic lacks a term for \textit{nephew} (expressed in Italian as \textit{nipotino}) \citep{Khalilia2023}. Such instances of \emph{lexical diversity}, referred to as \emph{(cross-lingual) lexical gaps} by \citet{giunchiglia2018one}, occur when a word in one language lacks a counterpart in another. When English is used as the reference language, language-specific concepts and lexical gaps may remain undocumented. Yet, cross-lingual NLP applications must account for phenomena of linguistic diversity \citep{giunchiglia2017understanding}. For example, machine translation systems often encounter lexical gaps. Google Translate and ChatGPT mistakenly render ``\textit{do not give cider to your child}'' into Arabic as \AR{لا تعطي عصير التفاح لطفلك}, which means ``\textit{do not give apple juice to your child}'', reflecting Arabic’s lack of a term for \textit{cider}. This highlights the challenge of achieving lexical equivalence across languages.

Addressing lexical diversity requires a \textit{systematic} approach to building diversity-aware datasets. To our knowledge, our expert-driven approach \citet{Khalilia2023} remains the only method that enables lexical gap identification at an advanced level within the target language, particularly in contexts where experts possess domain-specific knowledge. However, a major limitation of this approach is its \textit{unidirectional} design (English~$\rightarrow$~Target Language), which reinforces an \textit{English bias} and overlooks culture-specific concepts in non-English languages. Additionally, the reliance on professional linguistic experts significantly limits its applicability to \textit{low-resource languages}, thereby restricting coverage of global linguistic diversity.

Crowdsourcing has emerged as an effective means for developing NLP and linguistic resources, particularly those reflecting general language usage by native speakers. Prior efforts have included parallel corpus construction~\citep{post2012}, query systems like CrowdDB \citep{franklin2011}, WordNet development~\citep{ganbold2018}, lexicon enhancement~\citep{nair2022}, word sense disambiguation~\citep{parent2010}, sentiment analysis~\citep{kasumba2024}, and information retrieval~\citep{lease2012}.
In this paper, we aim to provide two key contributions:

\begin{enumerate}
\item A \textit{novel crowdsourcing methodology} for exploring lexical diversity across language pairs within specific semantic domains (e.g., food \citep{ashley2004}, kinship \citep{khishigsuren2022}, and body parts \citep{wierzbicka2007}). The method involves: (a)~semi-automated generation of lexical entries for each language, (b)~crowdsourcing evaluations by native speakers who compare lexical entries to identify meaning equivalents and lexical gaps, and (c)~validation by ordinary native speakers, followed by expert verification.
%
%
\item \textit{Empirical validation} of our method via two case studies involving English--Arabic and Indonesian--Banjarese language pairs, focused on food-related terminology. Across 132 tasks with 36~workers, we identified 2,140~lexical gaps in English--Arabic (1,532 in Arabic, 608 in English) and 951 (750 in Banjarese, 201 in Indonesian) in Indonesian--Banjarese, along with 1,957~equivalent terms.
\end{enumerate}

Our methodology is innovative in four key aspects:

\begin{enumerate}
    \item \textit{Language independence}: It applies to \textit{any language pair}, regardless of existing linguistic resources (e.g., lexical databases, encyclopedias, or digital and undigitized dictionaries or corpora).
    \item \textit{No reliance on pivot languages}: It does not depend on \textit{English} or any other language as an \textit{intermediary}.
    \item \textit{Bidirectional exploration}: It supports \textit{comparative} analysis from both source to target and vice versa.
    %
    %
    \item \textit{Applicability to both human and machine agents:} It can be implemented using either native-speaking crowd workers or large language models (LLMs). Our experiments show that native speakers are more effective than LLMs in identifying culturally and linguistically specific concepts, particularly in low-resource language contexts.
\end{enumerate}

The structure of the paper is as follows. In \Cref{secRelatedWork}, we review previous research related to our study. \Cref{Untranslatability} presents an overview of lexical diversity and lexical gaps. Our crowdsourcing methodology is described in \Cref{secMethod}, followed by its implementation and evaluation in \Cref{Sec_implementation_evaluation}. This includes the introduction of the LingoGap platform in \Cref{secLingoGap}, two case studies on food-related terminology in English–Arabic and Indonesian–Banjarese (\Cref{secArabicExp,secIndonesianExp}), and a comparison of crowdsourced data quality with LLM-generated annotations in \Cref{SecLLMs1}. In \Cref{secPersCrowd}, we discuss the use of crowdsourcing for constructing diversity-aware datasets. Finally, we conclude the paper in \Cref{secConclusion}.

\section{Related Work}
\label{secRelatedWork}

Crowdsourcing has been widely employed to create various linguistic resources, including lexical-semantic data. For instance, \citet{ganbold2018} developed a Mongolian WordNet using a two-phase crowdsourcing workflow via the CrowdCrafting\footnote{\url{https://crowdcrafting.org/}} platform. In the translation phase, volunteers suggested synonymous words by translating English PWN synsets into Mongolian. The subsequent validation phase employed inter-rater agreement metrics, such as Fleiss’ kappa and Krippendorff’s alpha, to ensure quality, achieving a precision of 0.74 for 947 synsets. Similarly, \citet{wijesiri2014} bootstrapped a Sinhala WordNet from English with the help of bilingual internet users. \citet{lanser2016} created a Japanese lexicon from DBpedia by first constructing an English version and then using annotators on CrowdFlower for translation. 

\citet{benjamin2014} introduced a mobile app–based crowdsourcing model ``\textit{Fidget Widget}'' to develop lexicons for low-resource languages. \citet{biemann2010} used Amazon Mechanical Turk (MTurk) to collect word senses for building a sense inventory from scratch. \citet{elhaj2015} recruited annotators to construct the Essex Arabic Summaries Corpus, yielding 2,360 sentences and 41,493 words in Jordanian and Gulf Arabic.

Other efforts include \citet{manerkar2022}, who developed ``\textit{Konkani Shabdarth}'', a crowdsourcing platform allowing community members (e.g., students and faculty members) to enhance to the Konkani WordNet by adding missing words to its synsets. \citet{fivser2014} introduced SloWCrowd to correct errors in the Slovene WordNet \cite{gantar2011}, while \citet{cibej2019} used crowdsourcing via PyBossa\footnote{\url{https://pybossa.com/}} to clean the Thesaurus of Modern Slovene. \citet{nair2022} proposed a Google Forms–based mobile approach for enhancing the Malayalam WordNet, referencing PWN.

The conventional method of \textit{expanding} WordNet through translation \citep{fellbaum2012} often fails to capture culture-specific concepts. For example, Arabic WordNet \citep{freihat2024} translates \textit{uncle} as \AR{عم} ``\textit{father’s brother}'', omitting maternal uncles. Bahasa WordNet \citep{noor2011} maps \textit{sister} to \textit{kakak} ``\textit{elder sibling}'', reducing semantic precision. In contrast, MultiWordNet \citep{pianta2002}, which employs a \textit{merge} strategy with bilingual dictionaries, explicitly captures lexical gaps but lacks coverage in rich semantic domains like kinship and food, and has since been discontinued.

Lexical typology, a subfield of linguistics, investigates cross-linguistic diversity by examining how languages encode meaning within specific semantic domains \citep{plungyan2011}. Lexical-typological research has explored translation-related challenges, particularly the presence or absence of lexicalized concepts across languages. Prior studies have focused on semantic domains known for considerable cross-linguistic variation, such as kinship terminology \citep{kemp2012}, color categories \citep{roberson2005}, food-related terms \citep{ashley2004}, human body parts \citep{wierzbicka2007}, and actions like cutting and breaking events \citep{majid2007} or putting and taking \citep{kopecka2012}. Despite ongoing research, publicly available datasets in this area remain scarce. Notable exceptions include \citet{murdock1970}’s kinship classification, which has been incorporated into D-PLACE \citep{kirby2016}, and aspects of \citet{kay2016}’s research on color terminology, available in the lexicon section of the World Atlas of Language Structures (WALS) \citep{dryer2013}. Another example is a dataset on color categorization by \citet{mccarthy2019}, accessible via GitHub\footnote{\url{https://github.com/aryamccarthy/basic-color-terms}}. 

Digital lexicons are increasingly employed in lexical typology, enabling researchers to analyze a broader range of languages and semantic domains. A notable example is the KinDiv\footnote{\url{http://ukc.disi.unitn.it/index.php/kinship}} lexicon \citep{khishigsuren2022}, which includes 1,911 lexical items and documents 37,370 lexical gaps related to kinship across 699 languages. Our study \citep{Khalilia2023} builds on this resource, specifically examining kinship-related lexical diversity in Arabic dialects and Indonesian languages. Other relevant research includes \citet{viberg1983}’s foundational study on perceptual vocabulary across 50 languages, which was later expanded by \citet{georgakopoulos2022} to incorporate data from 1,220 languages. 

Only one previous attempt to crowdsource lexical gaps is known—\citet{giunchiglia2015}, who developed a platform for translating English lexicalizations into Italian, including identifying lexical gaps, with the help of \emph{linguistic experts}. In contrast, our approach leverages \emph{non-expert native speakers} and differs significantly in its design and objectives.

Key differences between our approach and prior work include:

\begin{enumerate}
    \item \textit{No reliance on English as an intermediary:} Unlike methods that use English as a \textit{pivot}, our methodology compares datasets from any two languages directly, thereby avoiding \textit{English-centric bias}.
    \item \textit{Focus on lexical diversity and bidirectional exploration:} We exclusively target \textit{lexical diversity} and conduct \textit{bidirectional} comparisons to identify lexical gaps without assuming a fixed source-target direction. Contributions from native speakers are collected via micro-tasks enabled by our crowdsourcing platform.
\end{enumerate}

\section{Lexical Diversity and Cross-Lingual Lexical Gaps}
\label{Untranslatability}

Translation is a complex process influenced by cultural and lexical diversity, often resulting in challenges when striving for meaning equivalence across languages \citep{catford1965linguistic, bella2022}. Vocabulary embedded in specific cultural contexts—such as the Arabic terms for meals during Ramadan, a month in which Muslims fast from sunrise to sunset—demonstrates these challenges. For instance, \AR{السحور} \textit{suhur}, ``\textit{a pre-dawn meal consumed before the daily fast begins}'', and \AR{الافطار} \textit{iftar} ``\textit{the meal eaten at sunset to break the fast}'', encapsulate cultural practices that have no direct equivalents in many other languages. Similarly, culturally bound terms related to alcohol in European languages—such as the English \textit{Bitter}, ``\textit{a dry, sharp-tasting ale with a strong flavor of hops}'', the Italian \textit{Amaretto}, ``\textit{an almond liqueur}'', and the German \textit{Weizenbock} ``\textit{a wheat beer of bock strength}''. In Indonesia, rice is not merely a food item but a staple central to daily life and national identity. Indonesians use various terms to describe its forms: \textit{Gabah} ``\textit{harvested but unhulled rice}'', \textit{Beras} ``\textit{uncooked rice}'', \textit{Nasi} ``\textit{cooked rice}'', \textit{Kerak Nasi} ``\textit{scorched or crispy rice stuck to the bottom of the pot}''. These examples illustrate how languages encode distinct worldviews and culturally specific concepts that often resist direct translation.

In this study, we examine \textit{cross-lingual lexical gaps}—a phenomenon in which a word in the source language lacks a direct and precise equivalent in the target language. Such gaps often arise from cultural or regional specificities unique to individual linguistic communities and typically resist systematic translation through established rules or patterns \citep{lehrer1970}. A lexical gap is formally defined as follows:

\begin{definition}[Lexical Gap]
Let $L_1$ and $L_2$ be two natural languages, and let $w \in L_1$ be a lexical item expressing a well-defined meaning $m$. A lexical gap from $L_1$ to $L_2$ exists if there is no lexical item $w' \in L_2$ such that $w'$ conveys $m$ without semantic loss, approximation, or periphrasis.
\end{definition}

\vspace{-0.5\baselineskip}
To illustrate this, \Cref{tab:lex_foods_five_lang} presents examples of lexical gaps in food-related concepts across five languages. As the table shows, no single language offers concise lexicalizations for all the listed concepts, yet each concept is lexicalized in at least one language. These cross-linguistic variations pose challenges for both human and machine translation. Furthermore, substituting culturally specific terms with more general or approximate equivalents may lead to unintended meanings.

\begin{table}[]
    \centering
    \begin{tabular}{cccccc}
        \toprule
        \textit{Meaning} & \textbf{English} & \textbf{Japanese} & \textbf{Arabic} & \textbf{Italian} & \textbf{Indonesian} \\
        \midrule
         
        \textit{savory taste} & \cellcolor[HTML]{D3D3D3}GAP & \begin{CJK}{UTF8}{min}うま味\end{CJK} & \cellcolor[HTML]{D3D3D3}GAP & \cellcolor[HTML]{D3D3D3}GAP & gurih \\
         
        \textit{firm pasta texture} & \cellcolor[HTML]{D3D3D3}GAP & \cellcolor[HTML]{D3D3D3}GAP & \cellcolor[HTML]{D3D3D3}GAP & al dente & GAP \\
         
        \textit{water and dates} & \cellcolor[HTML]{D3D3D3}GAP & \cellcolor[HTML]{D3D3D3}GAP & \AR{الأسودان} & \cellcolor[HTML]{D3D3D3}GAP & \cellcolor[HTML]{D3D3D3}GAP \\
         
        \textit{crispy} & crispy & \cellcolor[HTML]{D3D3D3}GAP & \cellcolor[HTML]{D3D3D3}GAP & croccante & renyah \\
        \bottomrule
    \end{tabular}
    \caption{Lexicalizations of four meanings around the concept of (food) in five languages.}
    \label{tab:lex_foods_five_lang}
\end{table}


For example, the Arabic word \AR{الأسودان} ``\textit{water and dates}'' has no direct equivalent in English, as illustrated in our experiment in \Cref{secArabicExp}. This lexical gap can lead to mistranslations, such as Google Translate’s rendering\footnote{Translation retrieved on February 17, 2025} of the Arabic sentence \AR{يبدأ الصائم إفطاره بتناول الأسودين} as ``\textit{The fasting person begins his breakfast by eating lions}''. Another instance involves a mistranslation by ChatGPT-4 of the Indonesian word \textit{Kembili}—``\textit{a root vegetable similar to a potato}''—into Banjarese as \textit{Umbi-Umbian}, ``\textit{a broader term referring to various tuberous vegetables}''. Such inaccuracies highlight the need of identifying and addressing lexical gaps to preserve semantic integrity in cross-linguistic communication.

Recognizing patterns of lexical diversity highlights the need for scalable methods to document lexical gaps across a broad range of languages and semantic domains. However, existing approaches often depend on expert-driven processes that are \textit{English-centric} and \textit{unidirectional}—as discussed in \Cref{secRelatedWork}—which limits their scalability and accessibility, particularly for low-resource languages. To address these limitations, the following section introduces \textit{a novel crowdsourcing methodology} that leverages native speaker insights to systematically identify and verify lexical gaps.

\section{Crowdsourcing Methodology}
\label{secMethod}
This section outlines a methodology for the crowd-based collection of evidence related to lexical diversity. Lexical diversity is inherently an interlingual phenomenon, referring to meanings or distinctions that are not shared across languages. Such phenomena are especially common—but not limited to—socially or culturally significant domains, including food, religion, family relationships \citep{khishigsuren2022}, motion verbs \citep{walchli2012}, body parts \citep{wierzbicka2007}, colors \citep{mccarthy2019}, spatial dimensions \citep{Lang2001}, cutting and breaking events \citep{majid2007}, pain predicates \citep{reznikova2012}, perception verbs \citep{viberg1983}, and putting and taking events \citep{kopecka2012}. The \emph{initial inputs} to our methodology are, on the one hand, an \emph{ordered source--target language pair}, and on the other hand, a \emph{semantic field} (SF). The language pair is ordered because the methodology yields direction-dependent results: for the pair $(A,B)$, it identifies lexicalizations present in~$A$ but absent in~$B$, and for $(B,A)$, the reverse—lexicalizations present in~$B$ but not in~$A$. The \emph{output} of the method is a list of words in the source language (SL), each word annotated as either a lexical gap or lexicalized in the target language (TL), with an equivalent word provided in the latter case. This methodology defines two key roles: the task requester and the crowd worker. The \emph{task requester} is responsible for several critical functions: (1)~constructing the input datasets for both the SL and TL within the SF, (2)~designing and creating the crowdsourcing tasks, (3)~overseeing task execution to ensure high-quality contributions, and (4)~validating and exporting the finalized crowdsourced data. In contrast, the \emph{crowd worker} contributes by identifying equivalent terms and lexical gaps in the TL. The methodology is structured into three main steps:

\begin{enumerate}
\item \textit{Task generation}: A semi-automated process that produces two lists of \emph{lexical entries}, one for each language. A lexical entry is a tuple (\emph{word}, \emph{gloss})—that is, a term paired with its definition.
\item \textit{Crowdsourcing}: A \emph{crowdsourcing micro-task} consists of one lexical entry from the SL, along with all lexical entries from the TL as candidates. The goal is to determine whether the SL entry is translatable to the TL and, if so, to provide the equivalent TL entry. Our policy requires that crowd workers be native speakers of the TL and possess sufficient command of the SL. By “sufficient command”, we mean the ability to accurately understand the SL lexical entry, potentially with the aid of lexicons or online resources. In contrast, deciding whether an equivalent exists in the TL is a more challenging task that demands deep familiarity and active knowledge of the language—hence the requirement for TL-native workers.
\item \textit{Validation}: This step consolidates the contributions of crowd workers through a native speaker validation process, followed by expert-based verification.
\end{enumerate}

\subsection{Step 1: Task Generation}
\label{task_generation_step1}
Task generation takes lexical resources and corpora as input and produces a list of micro-tasks (i.e., \emph{word--gloss} tuples) for the workers. This process is semi-automatic: candidate micro-tasks are generated algorithmically and then filtered by an expert. The primary challenge lies in identifying candidate words that belong to the target semantic field in a manner that is robust across a wide range of languages and dialects, including low-resourced ones. We achieve this robustness by leveraging a variety of language resources, depending on what is available for the given language and semantic field. These resources may include mono- or multilingual lexical databases(e.g., wordnets), online dictionaries and encyclopedias (e.g., Wikipedia, Wiktionary, traditional dictionaries), language models or word embeddings, digital or undigitized corpora, and fieldwork with native speakers.

\textit{Lexical databases}, such as the UKC\footnote{\url{https://ukc.datascientia.eu}} or other wordnets, are preferred data sources due to their rich, structured, and meaning-annotated entries with definitions. Features such as domain tags and hierarchical relations support the filtering of entries by semantic field. However, only a few languages possess large-scale, high-quality lexical databases, and even fewer offer systematically annotated glosses.

\textit{Online dictionaries and encyclopedias} serve two purposes: (1)~providing glosses when they are absent from lexical databases—by extracting the first sentence from Wiktionary or Wikipedia for a given word, or discarding the word if no entry is found (in which case, no task is generated); and (2)~offering candidate words potentially related to the semantic field, which are later filtered using vector similarity methods.

\textit{Language models and word embeddings} are used to generate lexical entries when no adequate lexical resource exists in the source language for the given semantic field. A word list—obtained from a dictionary, as described above—is filtered to identify terms belonging to a specific semantic field (e.g., \emph{cake}, \emph{bread}, \emph{tomato} for \emph{food}). A vector representation of the field—constructed from a definition and example terms drawn from a small corpus—is generated using word or sentence embeddings (e.g., AraBERT for Arabic). These vectors are then compared to dictionary word vectors using cosine similarity, and the most similar words are selected.

\textit{Text corpora} can be used to train word embeddings or language models when pre-trained versions are not available for a given language. For languages without existing digital corpora, an initial corpus digitization step is required. 
Finally, \textit{fieldwork}, though often overlooked, is an effective method for obtaining a high-quality, focused corpus of words and definitions for a given language. It is particularly useful for low-resource languages and dialects when direct contact with native speakers is possible. The process begins with a language expert providing an initial set of seed words belonging to the semantic field. Native speakers then contribute additional words and definitions related to the field and the seed words.

For widely spoken languages such as English, Spanish, or Chinese, task generation is possible using a lexical DB alone (e.g., the UKC), although any combination of the aforementioned resources and methods may also be applied. For languages where lexical DBs are of lower quality, offer inadequate coverage, or do not exist—such as Arabic or Hungarian—data from lexical DBs can be supplemented with entries from traditional lexicons, filtered using language models. For even lower-resourced languages that lack language models but have usable corpora—such as European minority languages and dialects—digital dictionaries can provide candidate input words, and word embeddings can be trained and used to filter them according to the semantic field. Finally, for dialects and severely endangered languages with few or no existing corpora, results from prior fieldwork—such as kinship terms for Arabic dialects \citep{Khalilia2023}—can be used to produce smaller-scale but potentially high-quality word lists.

During the data preparation phase, SL terms are first collected from predefined semantic fields (e.g., Food, Emotion). For each SL entry, definitions are automatically extracted from multiple lexical resources such as Wiktionary, Wikipedia, and the UKC. When definitions are unavailable in these resources, alternative methods—such as corpus-based extraction, linguistic databases, or fieldwork—are employed to ensure comprehensive lexical coverage. As noted in Lines 233–235, \textit{any combination of the aforementioned resources and methods may be applied} to generate semantic field words and their corresponding definitions.

Once the SL definitions are established, TL candidate terms are retrieved semi-automatically from lexical databases using the same procedure applied to the SL terms. SL–TL pairs are then formed into tuples that serve as input for the subsequent crowdsourcing phase. Each tuple includes the SL word, its definition, and a list of TL candidates with their respective definitions.

For instance, within the Food semantic field, the English source term \textit{banana} is extracted from the SL lexical list together with its definition retrieved from Wiktionary. The initial TL candidates—such as \AR{موز} (Arabic)—are then automatically generated using an electronic dictionary. These preliminary SL–TL tuples are subsequently presented to annotators through the LingoGap interface, where participants validate the suggested equivalents, refine them as needed, or provide new TL terms if no appropriate match is found.


\subsection{Step 2: Crowdsourcing}
This section describes how the requester engages crowd workers to identify lexical gaps and equivalent terms between the SL and the TL. Crowd workers, recruited via a selected crowdsourcing platform, utilize the datasets created in the previous step. Since lexical diversity can arise in both the SL and the TL, the crowdsourcing process is conducted twice—once in each direction. In the first experiment, SL lexical entries—comprising word–gloss tuples—are mapped to the TL. In this phase, crowd workers identify equivalent terms as well as lexical gaps in the TL. The second experiment reverses the direction: the TL from the first experiment is treated as the new SL, and the original SL becomes the TL. Previously identified equivalents (i.e., overlapping terms) are excluded, and the focus shifts to mapping the remaining TL entries to the SL. Task crowdsourcing is organized into three phases, as described below:

 \begin{enumerate}
    \item \textit{Crowd selection}: The requester selects proficient crowd workers to participate in the task.
    \item \textit{Contribution collection}: Selected crowd workers complete micro-tasks on a crowdsourcing platform to provide equivalent terms and identify lexical gaps by comparing SL entries with TL entries, and vice versa.
    \item \textit{Contribution quality control}: The requester applies real-time quality control mechanisms to ensure the reliability and consistency of the collected data.
\end{enumerate}

\subsubsection{Crowd Selection} 
Ensuring high-quality responses from crowd workers is essential. We adopt a two-step selection process\footnote{Platforms like Prolific and MTurk offer built-in tools for quality control \citep{robinson2019}.}:

(1) \textit{Proficiency Test:} A preliminary test (comprising 10–25\% of the total questions) evaluates worker capability on the same platform as the main task \citep{liu2013}. To ensure domain-specific reliability, the test questions are contextualized within the semantic domain of the crowdsourcing task. Specifically, they assess workers’ understanding of domain-relevant vocabulary and concepts. For example, when the task involves the food domain, the test includes food-related terms; for a kinship domain, it focuses on kinship terminology. This domain-oriented design ensures that selected workers demonstrate both linguistic proficiency and domain-specific competence—or conceptual familiarity—necessary for producing high-quality annotations.\\

\begin{figure}[t!]
    \centering
    \includegraphics[width=\columnwidth]{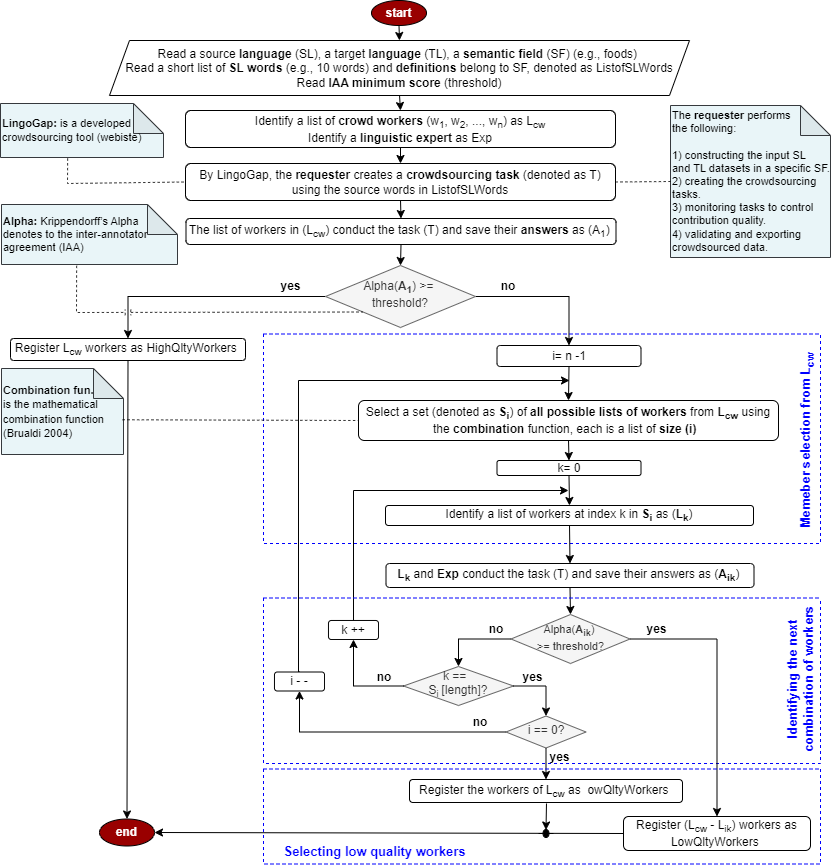} 
    \caption{Crowd Filtering using Alpha.}
    \label{fig2}
\end{figure}

(2) \textit{Filtering via Krippendorff’s Alpha:} To identify low-quality annotators, we compute inter-annotator agreement (IAA) using Alpha\footnote{Throughout this paper, the term ``Alpha'' refers specifically to Krippendorff’s Alpha''.}, which accounts for chance agreement, multiple annotators, and missing data. A subset of questions—contextualized within the food-related domain and comprising 25\% of the total items—is annotated by proficient workers and a linguistic expert to apply this filtering method (see \Cref{fig2}).


The worker filtering process is conducted on LingoGap—a custom-built crowdsourcing platform designed to collect lexical diversity data from non-expert native speakers—with input from a linguistic expert. The process begins by identifying the SL, TL, and SF, followed by the selection of approximately 10 SL terms within the chosen SF. After establishing an IAA threshold, workers are invited to complete a task on LingoGap. Workers whose agreement score (Alpha) meets or exceeds the threshold are classified as high-quality. For those who fall below the threshold, a subset of workers is evaluated—together with the expert—using all possible combinations of contributors (ranging from individual workers to all combinations except one), using the mathematical combination function described in \citet{brualdi2004}. Subsets that meet the IAA threshold are retained as high-quality, while excluded workers are categorized as low-quality.

\noindent\textbf{Example}

Consider a group of three workers, $G_1 = \{w_1, w_2, w_3\}$, performing a task. If their agreement (measured by Alpha) exceeds a predefined threshold, all are considered high-quality.

\noindent If not, an expert ($Exp$) helps identify the low-quality worker. The task is repeated with the expert and each pair of workers:
\vspace{-0.7em}
\[
\{Exp, w_1, w_2\}, \quad \{Exp, w_1, w_3\}, \quad \{Exp, w_2, w_3\}
\]
If any of these combinations meets the threshold, the excluded worker is flagged as low-quality.
If none meet the threshold, the task is repeated with the expert and each individual worker:
 \vspace{-0.7em}
\[
\{Exp, w_1\}, \quad \{Exp, w_2\}, \quad \{Exp, w_3\}
\]
If one combination passes, the other two workers are marked low-quality. If none pass, all workers in $G_1$ are classified as low-quality.

\subsubsection{Contribution Collection}
\label{Sec_ContCollection}
In this step, we use \textit{LingoGap}, described in \Cref{secLingoGap}, a crowdsourcing tool developed to identify lexical gaps and equivalent words in a given language pair.

A requester—using the \textit{admin} interface—creates a task and configures its details, including the description, language pair, and date. He or she selects source lemmas and their glosses from the SF dataset constructed in the source language. Additionally, He or she provides comprehensive \emph{instructions} and clear \emph{guidelines} for crowd workers through a customized spreadsheet template (see \Cref{tab:guidelines_english_arabic}), which includes nine default guidelines. These guidelines can be added, edited, or removed depending on the language pair involved in the experiment. For instance, one guideline prohibits the use of machine translation for defining words.

Once a task is created, crowd workers access the \textit{worker} interface\footnote{\url{http://lingogap.disi.unitn.it/}} and follow the provided guidelines to answer multi-step questions for each source word, presented sequentially. Each word prompts three \emph{multiple-choice questions (MCQs)}. The utility of MCQs for domain-targeted tasks has been demonstrated by \citet{welbl2017} in lexical semantic evaluations. The example below illustrates a semantic equivalence task for the English word “\textit{cider}” in comparison with Arabic:

\noindent
\textbf{Question:} ``\textit{Does the Arabic language include an equivalent meaning to the English word described below? If yes, please write the Arabic word along with its definition.}''  
\newline
\textbf{Word:} ``\textit{cider}''  
\textbf{Definition:} ``\textit{a beverage made from juice pressed from apples.}''

\begin{itemize}
    \item Choice 1: Yes, word: \AR{سايدر}, definition: \AR{مشروب كحولي مصنوع من التفاح}
    \item Choice 2: No
    \item Choice 3: Don’t know
\end{itemize}

Crowd workers can select one of three options: (1)~an equivalent meaning—Choice 1, retrieved from a precompiled Arabic food lexicon created during task setup; (2)~a lexical gap—Choice 2; or (3)~uncertainty—Choice 3, ``Don’t know''. Since the concept of ``\textit{cider}'' does not exist in Arabic, the correct answer is a lexical gap.

\subsubsection{Contribution Quality Control}
To ensure the reliability of crowdsourced data, we implement two live quality control mechanisms:

\noindent
(1) \textit{Attention Check Questions (ACQs).} ACQs are embedded within regular tasks to assess worker attentiveness and compliance with instructions. These simple questions are designed to detect careless responses. Following \citet{liu2013}, we include one ACQ for every ten questions and require workers to achieve at least 90\% accuracy, consistent with the threshold recommended by \citet{robinson2019}.

\noindent
(2) \textit{Completion Time Monitoring.} The time taken to complete a question serves as an indicator of engagement and response quality. Extremely short or long durations may suggest inattentiveness or rushed work. LingoGap logs completion times and automatically filters out outliers that deviate significantly from a worker’s average, ensuring only reliable data are retained.

\subsection{Step 3: Task Validation}
We validate the crowdsourced gaps and words in two subsequent phases. First, \textit{native-speaking crowd workers} perform data validation. We employ Alpha to measure IAA and filter out responses with low agreement. Second, a \textit{linguistic expert} reviews the responses with low IAA identified in the first phase.
 
\subsubsection{Crowd-Based Validation}
A group of proficient, native-speaking crowd workers—those involved in the contribution collection described in \Cref{Sec_ContCollection}—participate in a mutual validation process. In this process, each group cross-validates the contributions of another group, with the IAA scores across participants serving as the basis for evaluation.

IAA is widely used to assess the reliability of crowdsourced annotations in computational linguistics \citep{artstein2008}. Statistical measures such as Cohen’s Kappa \citep{warrens2011} and Krippendorff’s Alpha \citep{krippendorff2011} are commonly employed to evaluate consistency among annotators. Alpha is particularly versatile, accommodating various data types—nominal, ordinal, interval, and ratio—whereas Cohen’s Kappa is most appropriate for nominal data and pairwise agreement \citep{powers2012}.

In our crowdsourcing framework, which involves two or more annotators per item, we adopt Alpha to measure IAA. This approach enables us to systematically identify and exclude participants whose annotations lead to low agreement, ensuring that only items lacking sufficient consensus (with less than 100\% IAA) are escalated for expert review.

\begin{figure}[h!]
    \centering
    \includegraphics[width=0.92\columnwidth]{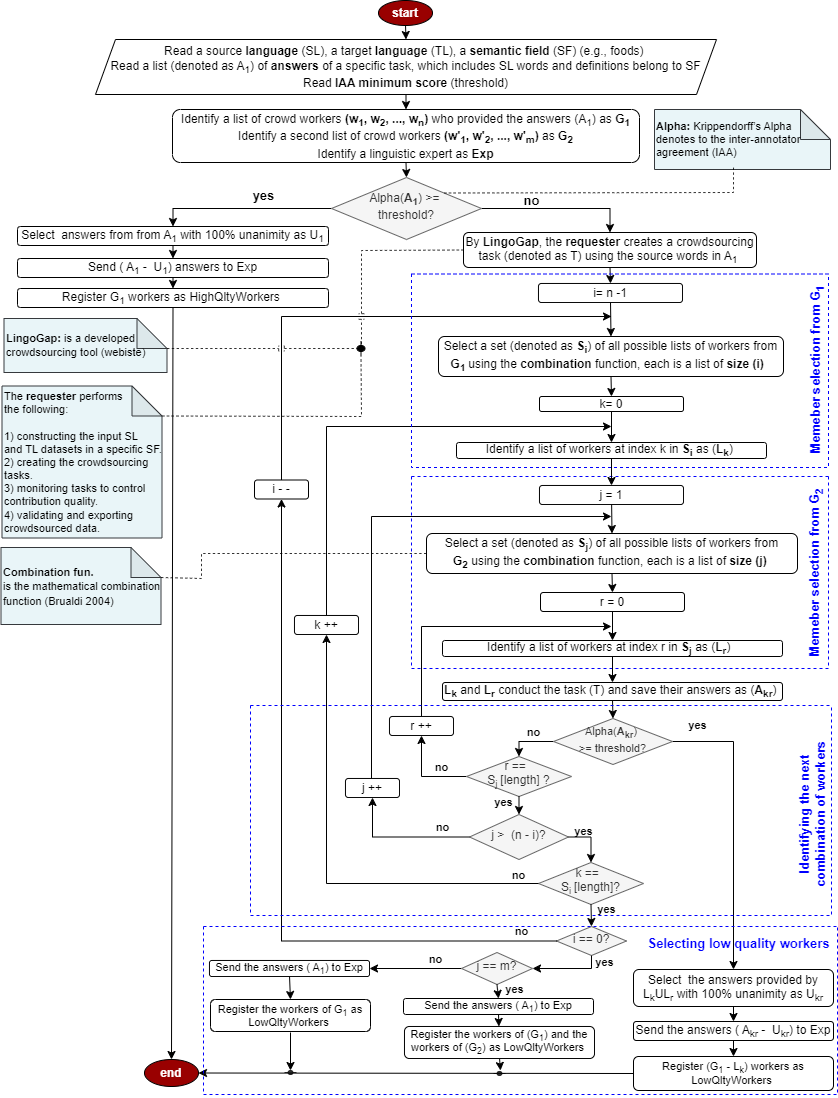} 
    \caption{Crowdsourced Data Validation using Alpha.}
    \label{fig3}
\end{figure}

To validate the data collected from an initial group of crowd workers ($G_{1}$: {$w_{1}$, $w_{2}$, $w_{3}$}), a second group of native-speaking crowd workers ($G_{2}$: {$w'_{1}$, $w'_{2}$, $w'_{3}$}) is engaged via the LingoGap platform. A visual overview of this procedure is provided in the flowchart in \Cref{fig3}. The validation process begins with reading the source and target language items, their definitions, and the responses collected from $G_{1}$. Inter-annotator agreement is then measured using Alpha. If the Alpha exceeds a predefined threshold, the $G_{1}$ workers are deemed high-quality, and only the items with disagreement are forwarded to a linguistic expert ($Exp$) for further evaluation.

If Alpha falls below the threshold, indicating low agreement, a new crowdsourcing task is launched through LingoGap. Subsets of workers from both $G_{1}$ and $G_{2}$ are formed using various combinations \citep{brualdi2004}, ranging from individual workers to the full group. These subsets repeat the annotation task, after which Alpha is recalculated. If a subset’s Alpha exceeds the threshold, the corresponding $G_{1}$ workers are classified as high-quality; those not included in such subsets are labeled as low-quality. An illustrative example is provided below.

%
\noindent \textbf{Example} \\
Consider two groups of crowd workers: $G_1 = \{w_1, w_2, w_3\}$ and $G_2 = \{w'_1, w'_2, w'_3\}$, along with a linguistic expert ($Exp$). 

First, the IAA is assessed by computing the Alpha value for $G_1$. If Alpha meets the required threshold, all $G_1$ workers are deemed high-quality, and any of their responses with less than 100\% agreement are sent to $Exp$ for validation.

If Alpha falls below the threshold, the process continues by forming combinations of two workers from $G_1$ with one from $G_2$, producing 9 groups (e.g., $\{w_1, w_2, w'_1\}, \{w_1, w_3, w'_1\}, \ldots$). If any combination meets the threshold, the excluded $G_1$ worker is marked low-quality, and their partial-agreement responses are sent to $Exp$.

If none pass, the process uses one worker from $G_1$ with two from $G_2$, forming another 9 combinations (e.g., $\{w_1, w'_1, w'_2\}, \{w_2, w'_1, w'_3\}, \ldots$). If any such combination exceeds the threshold, the remaining $G_1$ workers are marked low-quality, and their responses with less than 100\% IAA are sent to $Exp$.

If all these also fail, the task is repeated using only $G_2$. If their Alpha meets the threshold, all $G_1$ workers are considered low-quality. Otherwise, both groups are marked low-quality, and all $G_1$ responses are sent to $Exp$, concluding the validation.

\subsubsection{Expert-based Verification} 
A bilingual expert reviews responses from the Crowd-Based Validation stage that did not achieve 100\% IAA. A spreadsheet is prepared containing the following columns: \emph{Worker-ID}, \emph{Source Lemma}, \emph{Source Gloss}, and \emph{Worker’s Answer} (categorized as ``\textit{Lexical Gap}'', ``\textit{Equivalent Word}'', or ``\textit{Do not know}'').

Additionally, 10\% of the entries in the spreadsheet—randomly selected from those with 100\% IAA—are included as a ``sanity check" to assess the expert’s consistency. The expert is then tasked with the following:

\begin{itemize}
    \item \textit{Equivalent Meanings}: Evaluate the target language words provided by crowd workers and mark them as correct or incorrect. If a word is deemed incorrect, the expert supplies the correct equivalent or identifies it as a lexical gap.
    \item \textit{Lexical Gaps}: Confirm or reject the crowd’s classification of source words as lexical gaps. If a word is incorrectly marked as a gap, the expert provides the appropriate equivalent.
    \item \textit{Do Not Know}: For source words marked as ``\textit{Do not know}'', the expert determines whether they represent lexical gaps or provides suitable equivalents in the target language.
\end{itemize}

\section{Implementation and Evaluation}
\label{Sec_implementation_evaluation}
In this section, we present the practical implementation and evaluation of our crowdsourcing methodology via the LingoGap platform. We begin by outlining the platform’s technical and functional features. Subsequently, we describe two case studies within the food semantic domain, involving diverse language pairs—English–Arabic and Indonesian–Banjarese—selected for their cultural and lexical variation. Finally, we evaluate the quality of the crowdsourced data by comparing it with annotations produced by large language models (LLMs), offering insights into the respective strengths and limitations of human and machine contributions in identifying lexical gaps.

\subsection{LingoGap Platform}
\label{secLingoGap}

To operationalize our crowdsourcing methodology, we developed LingoGap, a custom-built, web-based platform for collecting, managing, and validating lexical diversity data through structured micro-tasks. Developed using Java (JSP), JavaScript, and MySQL, LingoGap supports both administrative (requester) and worker roles, and it can function as a standalone system or integrate with commercial crowdsourcing platforms such as Prolific. The platform features two primary interfaces: the \textit{requester interface} for researchers and the \textit{worker interface} for crowd contributors.

\textbf{Requester Interface\footnote{\url{http://lingogap.disi.unitn.it/admin.jsp}}}: The interface includes multiple tabs for task management. Using the ``\textit{Experiments}'' tab, the requester can import SF datasets created in \Cref{task_generation_step1}, configure tasks by specifying the SL and TL, and select a subset of words (e.g., 35~terms) from the full SL dataset displayed in a data table. The requester can also monitor task execution by defining ACQs and recording the completion times of crowd workers’ responses. Through the ``\textit{Source Words}'' and ``\textit{Target Words}'' tabs, lexical entries for both languages can be loaded and managed. Additionally, the requester can upload customized task instructions to enhance clarity. Requesters may import guideline spreadsheets into the LingoGap database. These spreadsheets contain configurable prompts (e.g., explanations of lexical gaps, task restrictions), expected response formats, and ethical considerations (e.g., anonymity, withdrawal rights), enabling the creation of flexible instruction templates tailored to the specific source and target languages. For instance, the \emph{guidelines} used in our English-Arabic case study (\Cref{secArabicExp}) are detailed in \Cref{tab:guidelines_english_arabic}.

\begin{table}[htbp]
    \centering
    \small 
    \begin{tabular}{m{0.19\linewidth}m{0.8\linewidth}} 
        \hline
        What will you be asked to do? & You will be asked to inspect 35 English food words and evaluate them by selecting one of two alternatives for each word: whether Arabic has an equivalent meaning or it is a lexical gap. \\
        \hline
        What is a lexical gap? & A lexical gap is a word with a distinct meaning that is missing from the vocabulary of a language. \\
        \hline
        What is an example of a lexical gap? & The English term "ham sandwich," referring to a sandwich filled with sliced ham, is a lexical gap in Arabic due to cultural differences. In many Arabic-speaking cultures, ham is not commonly consumed because of cultural and religious considerations, and sandwiches containing ham are not typically found in Arabic cuisine. Similarly, the Arabic word \AR{سحور} meaning "the meal that Muslims eat before dawn during the month of Ramadan," represents a gap in the English-speaking community. \\
        \hline
        What are the needed qualifications? & Our experiment is not restricted by the user's background, culture, skills, etc. The only requirement is that participants must be native Arabic speakers with a good command of the English language. We prefer participants to have linguistic knowledge. \\
        \hline
        Are there any restrictions? & The only restriction is that you are not allowed to use machine translation, such as Google Translate, for translating word definitions. All possible meanings in Arabic are presented in a list in the answer section. \\
        \hline
        How long will the experiment take? & The maximum duration of the experiment is about 60 minutes. Please try to be as accurate as possible. \\
        \hline
        Tips: & We encourage you to (1) use a dictionary if you are unsure about your answer, and (2) search for an image of the English word on Wikipedia if you do not understand the English definition. \\
        \hline
        Will your data be processed anonymously? & The data collected will be kept strictly confidential; all responses will be stored and processed anonymously. \\
        \hline
        How will you indicate your consent? & By clicking "I consent, begin the study" below, you acknowledge that you speak English and Arabic, that you are at least 18 years old, and that you give your consent to participate in this study. If you do not intend to give consent, click "I do not consent; I want to withdraw from this study." Even after giving consent, you have the right to stop and withdraw from the experiment at any time without providing a reason. You can withdraw from the study at any time simply by closing the browser. \\
        \hline
        Contact person & If you have any questions, please contact the task requester (Hadi Khalilia) via email at hadi.khalilia@unitn.it. \\
        \hline
    \end{tabular}
    \caption{Guidelines for the experiment of English to Arabic  (described in \Cref{secArabicExp}).}
    \label{tab:guidelines_english_arabic}
\end{table}

\begin{figure}[htbp]
    \centering
    \includegraphics[width=\textwidth]{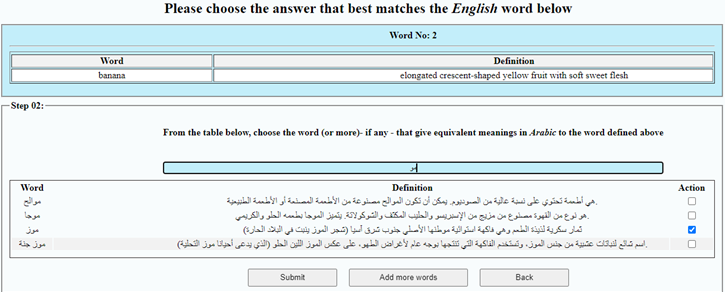}
    \caption{Worker’s GUI showing the step of match selection using LingoGap.}
    \label{fig4_workerGUI}
\end{figure}

\textbf{Worker Interface}: Workers are presented with SL words and complete a multi-step evaluation for each word: (1)~Gap Assessment – Determine whether the meaning exists in the TL. If not, the item is marked as a lexical gap; (2)~Match Selection – If the meaning exists, select the equivalent from a provided list of TL words. For instance, \Cref{fig4_workerGUI} shows a list of Arabic words for the selection of a food-related term in the example of exploring ``\textit{banana}'' in \Cref{secArabicExp}; and (3)~Custom Entry – If no suitable match is found, manually input a TL term along with a gloss. This multi-step format, \textit{conceptually} inspired by the structured reasoning principle of Chain-of-Thought (CoT) approaches \cite{lin2024}, encourages annotators to follow a clear, stepwise decision process that enhances task comprehension.


\subsection{Case Study on Food Terminology Across English and Arabic}
\label{secArabicExp}

This case study investigates lexical diversity in the food domain across English and Arabic using the crowdsourcing methodology described in \Cref{secMethod}. The study was conducted in two phases: English-to-Arabic and Arabic-to-English, following the same structured workflow.

\subsubsection{Study Setup}

\begin{figure}[t!]
    \centering
    \includegraphics[width=\linewidth]{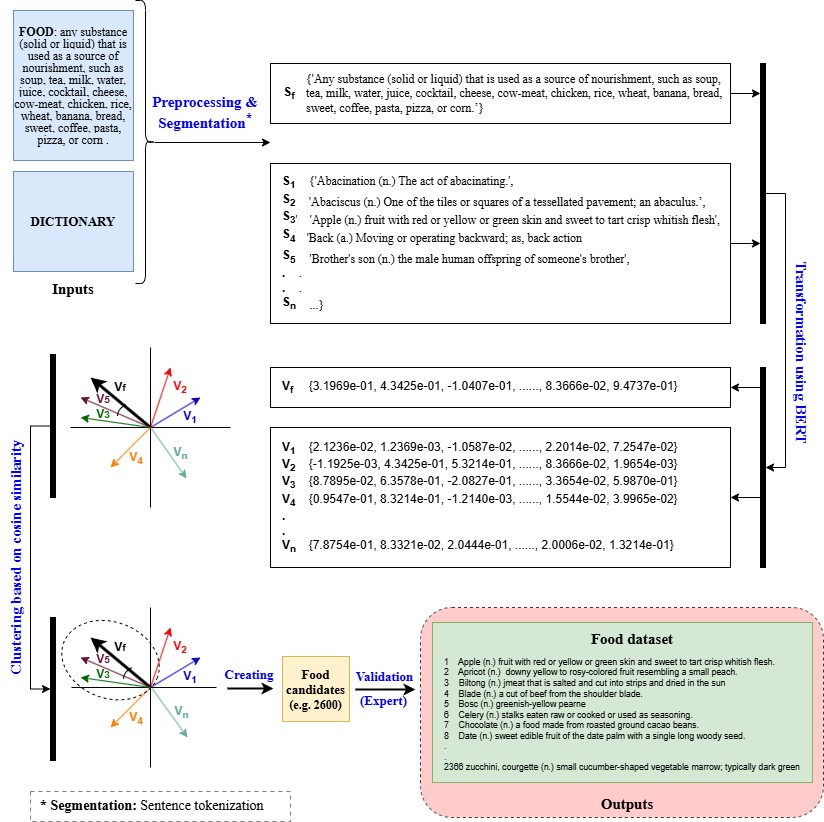}
    \caption{Flowchart of the semantic-field filtering method to collect food words from a digital dictionary.}
    \label{fig:ArabicDSCreation}
\end{figure}

\noindent
\textbf{Task generation.} 
Two lexical datasets were developed: one in English (2,364~terms) and the other in Arabic (1,607~terms). The English food terms were extracted from the UKC using the Exporter tool, a built-in UKC management service designed to extract data from the UKC. The tool was used to collect English words belonging to a specific semantic field. It takes one or more concepts (e.g., food, nutrient) that represent a given semantic field (e.g., food) and the UKC resource as inputs, and then retrieves all lexicalizations associated with those concepts from the corresponding language lexicon in the UKC.

For Arabic, due to the lack of comprehensive lexical databases—particularly those containing food-related terms (e.g., UKC, WordNet)—a custom semantic filtering approach was employed. This method, based on AraBERT embeddings, extracted food-relevant terms from digital dictionaries (e.g., Almaany Dictionary) using cosine similarity. As illustrated in \Cref{fig:ArabicDSCreation}, the method involves three main steps. In \emph{Step~1}, a digital dictionary and food-related definitions (including example terms) are provided as inputs, which are then preprocessed by segmenting paragraphs into sentences. In \emph{Step~2}, the sentences are transformed into vector representations using AraBERT. In \emph{Step~3}, dictionary vectors similar to the centroid food vector are clustered based on cosine similarity with an 0.85 threshold, then converted back into text sentences and compiled in a spreadsheet.
For clarity and ease of understanding, \emph{English} examples were used for the input and output demonstrations instead of Arabic ones. A custom Python script implementing this AraBERT-based methodology was developed to collect Arabic food-related words from the dictionary. The script is available on GitHub\footnote{\url{https://github.com/HadiPTUK/developed_scripts/blob/main/arabic_food_terms_script.py}}.



\noindent
\textbf{Crowd selection.}
Native speakers proficient in the target language were recruited as volunteers: 12 Arabic native speakers with English fluency and 12 English native speakers with Arabic proficiency. All participants were university-educated. Each group was divided into four teams (G1–G4) of three members. A proficiency test and Alpha–based filtering methodology (depicted in \Cref{fig2}) were applied to identify high-quality contributors. For instance, 12 of 14 students passed the test successfully in the experiment of (English → Arabic), and for the application the filtering methodology in the same experiment, four tasks-each comprising 10~questions-were conducted . $G_{1}$ completed the first task, $G_{2}$ the second, $G_{3}$ the third, and $G_{4}$ the fourth, with all tasks performed by a linguistic expert. As a result, one student from $G_{1}$ was replaced, while $G_{2}$, $G_{3}$, and $G_{4}$ remained unchanged.

\noindent
\textbf{Contribution Collection and Quality Control.}
Using the LingoGap platform and the guidelines outlined in \Cref{tab:guidelines_english_arabic}, each group completed micro-tasks to evaluate source language terms and determine whether equivalent terms existed in the target language or whether a lexical gap was present. Three response options were available: equivalent term, lexical gap, or ``do not know''. Tasks included automated logging of completion times, ACQs to ensure attentiveness, and the exclusion of responses with anomalously short or long completion times. For example, in the \textit{English → Arabic} experiment, for each English word, LingoGap displayed the term and asked crowd workers whether an equivalent existed in Arabic. If confirmed, LingoGap presented a list of Arabic food-related terms for selection, as illustrated with the term \textit{banana} in \Cref{fig4_workerGUI}. If no equivalent was confirmed, the response was recorded as a lexical gap. If the equivalent term was not listed, the crowd was allowed to enter it manually via text input.

In the \textit{English → Arabic} setting, each group processed subsets of the 2,364~English words against the full Arabic dataset. Groups $G_1$ and $G_2$ each worked on 245~English words distributed across seven crowdsourcing tasks covering \textit{alcoholic drinks, pizza, salads, dairy products, rice, bread}, and \textit{fruits}. Groups $G_3$ and $G_4$ worked on 945 and 929~English words, respectively, focusing on \textit{soups, vegetables, cakes, meats, sandwiches}, and \textit{desserts}.

In the \textit{Arabic → English} direction, overlapping terms identified as equivalents in the (English → Arabic) experiment were excluded, resulting in 906~unique Arabic terms. The groups worked with these filtered Arabic terms evaluated against the full English dataset. Group $G_1$ was assigned 245~Arabic words and $G_2$ 241~words, with both groups covering tasks on \textit{sweets, rice meals, soups, vegetables, meats}, and \textit{sandwiches}. Groups $G_3$ and $G_4$ each addressed 210~Arabic words across six tasks focusing on \textit{drinks, pizza, salads, dairy products, bread}, and \textit{fruits}.

Pilot tasks were conducted to refine the task design. Four pilot tasks were used to calibrate task parameters including the number of questions, number of ACQs, and task duration. The final configuration was set to 35~questions, 3~ACQs, and 60 minutes per task. Based on pilot task outcomes, three ACQs were adopted in all subsequent tasks. For instance, in Task~6 of the \textit{English → Arabic} experiment conducted by $G_2$, responses from worker $w^{\prime}_{1}$ were excluded due to a failed ACQ. Responses from $w^{\prime}_{2}$ and $w^{\prime}_{3}$, with an Alpha of 0.922, were retained (see Table~\ref{table4_g1_g2_new}). Additionally, outlier responses were filtered based on completion time per word. Among 68~tasks, 19~responses were discarded due to unusually fast completion times (3–6~seconds), compared to the average range of 80–120~seconds per task.

\noindent
\textbf{Task Validation.}
For both the \textit{English → Arabic} and \textit{Arabic → English} experiments, we employed the methodology illustrated in \Cref{fig3} to validate crowdsourced data from groups $G_{1}$, $G_{2}$, $G_{3}$, and $G_{4}$. IAA, measured using Alpha, generally exceeded the acceptance threshold of 0.70. However, in the \textit{English → Arabic} setting, two exceptions were noted: Task~5 (by $G_{2}$) and Task~17 (by $G_{3}$) initially scored Alpha values of 0.59 and 0.62, respectively. These tasks were subsequently reassigned—Task~5 to workers from $G_{1}$ and $G_{2}$, and Task~17 to workers from $G_{3}$ and $G_{4}$—resulting in improved Alpha scores of 0.89 and 0.82. Workers $w^{\prime}_{3}$ (from $G_{2}$) and $w^{\prime}_{1}$ (from $G_{3}$) were replaced in later tasks due to inaccuracies that contributed to disagreement.

In contrast, all tasks in the \textit{Arabic → English} experiment achieved Alpha values above 0.70, as shown in \Cref{tab:crowdsourced_data_english:g1_and_g2,tab:crowdsourced_data_english:g3_and_g4}. Despite satisfactory agreement levels, all newly introduced words and responses with less than 100\% IAA were submitted for expert validation in both experiments. In the \textit{English → Arabic} direction, this included 88~English words. The expert confirmed the validity of all proposed Arabic words and suggested alternatives for two lexical gaps. For example, \AR{خبز شراك} ``\textit{khubz shrak}'' was recommended for “\textit{chapatti}” instead of leaving a gap, as initially proposed by $G_{3}$ in Task~12.

Similarly, in the \textit{Arabic → English} experiment, the expert reviewed 79~Arabic words with non-unanimous responses. All new English translations were validated, though four Arabic words were identified as lexical gaps. For instance, \AR{مهلبية}—``\textit{a dessert made from milk, starch, and sugar}''—was classified as a lexical gap, with the previously suggested equivalent ``\textit{pudding}'' (provided by $G_{2}$ in Task~4) deemed too broad in semantic scope.

\subsubsection{Study Results}
\label{ArabicResults}

This section presents the results of two crowdsourcing experiments-English-to-Arabic and Arabic-to-English mappings- conducted to explore lexical diversity between English and Arabic in the domain of food-related terms. 

Across the \textit{English → Arabic} experiment, participants identified 1,532 lexical gaps, 832 equivalent words, and 100 new Arabic words that were not present in the original Arabic input dataset. The annotation process achieved a high inter-annotator agreement, with an average Alpha score of 0.84. Detailed information for the Arabic crowdsourced data are presented in \Cref{table4_g1_g2_new,table4_NEW}.

%

\begin{table}[h]
\caption{Crowdsourced data summary by $G_{1}$ and $G_{2}$ in Arabic.}
\label{table4_g1_g2_new}
\centering
\begin{tabular}{ccccccccc}
\toprule
\textbf{Task} & \multicolumn{2}{c}{\textbf{Gaps}} & \multicolumn{2}{c}{\textbf{Words}} & \multicolumn{2}{c}{\textbf{New Concepts}} & \multicolumn{2}{c}{\textbf{Alpha}} \\
& G1 & G2 & G1 & G2 & G1 & G2 & G1 & G2 \\
\midrule
1 & 19 & 32 & 16 & 3 & 1 & 2 & 0.76 & 0.72 \\
2 & 34 & 30 & 1 & 5 & 0 & 1 & 1.00 & 0.73 \\
3 & 34 & 26 & 1 & 9 & 3 & 1 & 1.00 & 0.95 \\
4 & 19 & 29 & 16 & 6 & 2 & 1 & 0.80 & 0.72 \\
5 & 27 & 21 & 8 & 14 & 2 & 0 & 0.71 & 0.89 \\
6 & 26 & 21 & 9 & 14 & 2 & 3 & 0.85 & 0.92 \\
7 & 21 & 26 & 14 & 9 & 3 & 3 & 0.92 & 0.77 \\
\midrule
\textbf{Total} & 180 & 185 & 65 & 60 & 13 & 11 & & \\
\textbf{Average} & & & & & & & 0.86 & 0.81 \\
\bottomrule
\end{tabular}
\end{table}

%

\begin{table}[h]
\caption{Crowdsourced data summary by $G_{3}$ and $G_{4}$ in Arabic.}
\label{table4_NEW}
\centering
\begin{tabular}{cccccccccc}
\toprule
\textbf{Task} & \multicolumn{2}{c}{\textbf{Gaps}} & \multicolumn{2}{c}{\textbf{Words}} & \multicolumn{2}{c}{\textbf{New Concepts}} & \multicolumn{2}{c}{\textbf{Alpha}} \\
 & G3 & G4 & G3 & G4 & G3 & G4 & G3 & G4 \\
\midrule
1 & 29 & 23 & 6 & 12 & 1 & 1 & 0.81 & 0.88 \\
2 & 21 & 33 & 14 & 2 & 2 & 2 & 0.73 & 0.79 \\
3 & 24 & 20 & 11 & 15 & 2 & 1 & 0.79 & 0.92 \\
4 & 17 & 21 & 18 & 14 & 1 & 2 & 0.92 & 0.91 \\
5 & 19 & 15 & 16 & 20 & 2 & 2 & 0.81 & 0.80 \\
6 & 24 & 27 & 11 & 8 & 0 & 0 & 0.83 & 0.95 \\
7 & 18 & 22 & 17 & 13 & 0 & 0 & 0.79 & 0.88 \\
8 & 24 & 20 & 11 & 15 & 3 & 1 & 0.77 & 0.87 \\
9 & 24 & 13 & 11 & 22 & 2 & 2 & 0.71 & 0.92 \\
10 & 14 & 24 & 21 & 11 & 2 & 0 & 0.76 & 0.96 \\
11 & 20 & 21 & 15 & 14 & 0 & 2 & 0.87 & 0.89 \\
12 & 25 & 16 & 10 & 19 & 2 & 2 & 0.78 & 0.88 \\
13 & 29 & 23 & 6 & 12 & 1 & 2 & 0.93 & 0.84 \\
14 & 30 & 26 & 5 & 9 & 0 & 1 & 0.86 & 0.91 \\
15 & 23 & 27 & 12 & 8 & 2 & 2 & 0.92 & 0.90 \\
16 & 32 & 28 & 3 & 7 & 0 & 1 & 0.78 & 0.79 \\
17 & 18 & 16 & 17 & 19 & 1 & 3 & 0.62 & 0.89 \\
18 & 24 & 22 & 11 & 13 & 2 & 1 & 0.91 & 0.88 \\
19 & 14 & 19 & 21 & 16 & 1 & 2 & 0.80 & 0.92 \\
20 & 22 & 21 & 13 & 14 & 2 & 3 & 0.87 & 0.73 \\
21 & 19 & 22 & 16 & 13 & 1 & 0 & 0.89 & 0.88 \\
22 & 24 & 23 & 11 & 12 & 1 & 2 & 0.91 & 0.72 \\
23 & 17 & 20 & 18 & 15 & 0 & 3 & 0.81 & 0.81 \\
24 & 19 & 21 & 16 & 14 & 1 & 3 & 0.85 & 0.85 \\
25 & 24 & 24 & 11 & 11 & 3 & 2 & 0.87 & 0.87 \\
26 & 23 & 22 & 12 & 13 & 0 & 0 & 0.90 & 0.84 \\
27 & 15 & 6 & 20 & 13 & 3 & 1 & 0.83 & 0.86 \\
\midrule
\textbf{Total} & 592 & 575 & 353 & 354 & 35 & 41 & \multicolumn{2}{c}{} \\
\textbf{Average} & \multicolumn{6}{c}{} & 0.83 & 0.86 \\
\bottomrule
\end{tabular}
\end{table}

%


Conversely, in the \textit{Arabic → English} direction, the experiment yielded 608~lexical gaps, 298~equivalent words, and 49~new English words not found in the original English input. This task also demonstrated strong annotator consistency, with an average Alpha score of 0.85. Additional information on the English crowdsourced data is provided in \Cref{tab:crowdsourced_data_english:g1_and_g2,tab:crowdsourced_data_english:g3_and_g4}.


\begin{table}[h]
    \centering
    \caption{Crowdsourced data summary by $G_{1}$ and $G_{2}$ in English.}
    \label{tab:crowdsourced_data_english:g1_and_g2}
    \begin{tabular}{ccccccccc}
        \toprule
        \textbf{Task} & \multicolumn{2}{c}{\textbf{Gaps}} & \multicolumn{2}{c}{\textbf{Words}} & \multicolumn{2}{c}{\textbf{New Concepts}} & \multicolumn{2}{c}{\textbf{Alpha}} \\
        & G1 & G2 & G1 & G2 & G1 & G2 & G1 & G2 \\
        \midrule
        1 & 29 & 19 &  6 & 16 & 3 & 2 & 0.87 & 0.92 \\
        2 & 25 & 24 & 10 & 11 & 1 & 0 & 0.75 & 0.92 \\
        3 & 24 & 19 & 11 & 16 & 2 & 1 & 0.82 & 0.85 \\
        4 & 27 & 28 &  8 &  7 & 2 & 2 & 0.84 & 0.80 \\
        5 & 23 & 25 & 12 & 10 & 1 & 2 & 0.89 & 0.81 \\
        6 & 27 & 19 &  8 & 16 & 2 & 2 & 0.83 & 0.77 \\
        7 & 21 & 29 & 14 &  2 & 1 & 3 & 0.77 & 0.85 \\
        \midrule
        \textbf{Total} & 176 & 163 & 69 & 78 & 12 & 12 & \multicolumn{2}{c}{} \\
        \textbf{Average} & \multicolumn{6}{c}{} & 0.82 & 0.85 \\
        \bottomrule
    \end{tabular}
\end{table}

\begin{table}[h]
    \centering
    \caption{Crowdsourced data summary by $G_{3}$ and $G_{4}$ in English.}
    \label{tab:crowdsourced_data_english:g3_and_g4}
    \begin{tabular}{ccccccccc}
        \toprule
        \textbf{Task} & \multicolumn{2}{c}{\textbf{Gaps}} & \multicolumn{2}{c}{\textbf{Words}} & \multicolumn{2}{c}{\textbf{New Concepts}} & \multicolumn{2}{c}{\textbf{Alpha}} \\
        & G3 & G4 & G3 & G4 & G3 & G4 & G3 & G4 \\
        \midrule
        1 & 21 & 26 & 14 &  9 & 0 & 2 & 0.81 & 0.91 \\
        2 & 18 & 26 & 17 &  9 & 2 & 3 & 0.89 & 0.95 \\
        3 & 32 & 22 &  3 & 13 & 3 & 1 & 0.72 & 0.88 \\
        4 & 15 & 24 & 20 & 11 & 3 & 2 & 0.88 & 0.87 \\
        5 & 21 & 22 & 14 & 13 & 4 & 2 & 0.87 & 0.85 \\
        6 & 15 & 27 & 20 &  8 & 2 & 1 & 0.92 & 0.81 \\
        \midrule
        \textbf{Total} & 122 & 147 & 88 & 63 & 14 & 11 & \multicolumn{2}{c}{} \\
        \textbf{Average} & \multicolumn{6}{c}{} & 0.85 & 0.88 \\
        \bottomrule
    \end{tabular}
\end{table}

The resulting datasets are publicly available via the DataScientia repository. The \textit{English → Arabic} dataset can be accessed at \footnote{\url{https://ds.datascientia.eu/community/public/projects/aef86b5b-5742-4529-9c46-71b5142f91bd}}, and the \textit{Arabic → English} dataset is available at \footnote{\url{https://ds.datascientia.eu/community/public/projects/db9ec453-4775-43d6-a89a-85a9600b9781}}.

\subsubsection{Lexical Diversity Evaluation: Overlap-Based Metric}

Several shared meaning overlaps have been found between language pairs. For a given domain $d$ and two languages $l_A$ and $l_B$, the formula below calculates the similarity of the two languages in terms of the overlap of lexicalised concepts from that domain, where $\textrm{LexCons}(d, l)$ stands for the set of domain concepts that are lexicalized by the language~$l$.

\begin{equation}
    \textrm{overlap}(d, l_A,l_B) = \frac{|\textrm{LexCons}(d, l_A)\cap \textrm{LexCons}(d, l_B)|}{\textrm{max}(|\textrm{LexCons}(d, l_A)|, |\textrm{LexCons}(d, l_B)|)} 
    \label{eq:crowdsourcing_overlap}
\end{equation}

\Cref{fig:venn_eng_arb} shows the overlaps between English and Arabic over the food domain. For example, the intersection of English and Arabic languages gives a shared coverage of 46.8\%. The number of lexicalisations in English is 2413~(2364~words in the English input dataset and 49~new words, which are missing from the English input dataset), and in Arabic is 1707~(1607~words in the Arabic input dataset and 100~new words). Also, 1130 of these lexical units (832~words were explored in the experiment of \textit{English → Arabic}, 298~words were identified in the experiment of \textit{Arabic → English}) are included in both languages. For example, \Cref{eq:crowdsourcing_overlap} calculates the overlap between English and Arabic (both represented in ISO 639-3 as ``eng'' and ``arb'', respectively) in the food domain ($F$) as follows:

\begin{figure}[h]
    \centering
    \includegraphics[width=0.5\columnwidth]{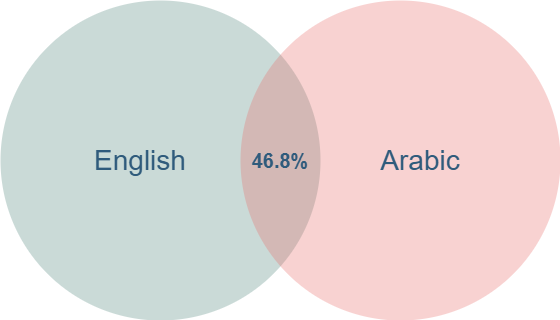} 
    \caption{The overlap (percentage of shared lexicalizations) for English and Arabic languages.}
    \label{fig:venn_eng_arb}
\end{figure}

\begin{equation}
    \mathrm{overlap}(F, \mathrm{eng}, \mathrm{arb})
    =
    \frac{|\mathrm{LexCons}(F, \mathrm{eng})\cap \mathrm{LexCons}(F, \mathrm{arb})|}{\textrm{max}(|\mathrm{LexCons}(F, \mathrm{eng})|, |\mathrm{LexCons}(F, \mathrm{arb})|)}
\end{equation}

\begin{equation}
    \mathrm{overlap}(F, \mathrm{eng}, \mathrm{arb})=\frac{1130}{\mathrm{max}(2413, 1707)}=\frac{1130}{2413}=46.8\%
\end{equation}

We find this overlap is lower than our initial expectations on language variations. Language experts justify such differences with two major factors: linguistic and religious influence \citep{albala2011,armanios2018}. By linguistic influence, we refer to the etymological origin and borrowing of the language, which affects the lexicons. The two languages have distinct etymological origins. Arabic, a Semitic language, derives many food terms from ancient roots and influences from Persian, Turkish, and Indian cultures. English, a Germanic language, has borrowed food-related vocabulary from French, Italian, and other European languages over time. Secondly, the religion of the speaker community also affects the lexicon. In many Arabic-speaking regions, Islam significantly influences food practices. Halal dietary laws shape food culture, while pork and alcohol, common in some Western cuisines, are prohibited in the Arabic community. English-speaking countries have more religious diversity, which can allow for a broader range of food-related terms tied to various cuisines \citep{armanios2018}.

\subsection{Case Study on Food Terminology Across Indonesian and Banjarese}
\label{secIndonesianExp}

To investigate lexical diversity in the food domain between Standard Indonesian\footnote{Throughout this paper, the term ``Indonesian'' refers specifically to Standard Indonesian''.} and Banjarese, we conducted two bidirectional experiments—matching from Indonesian to Banjarese and vice versa—using the crowdsourcing methodology described in \Cref{secMethod}.


\subsubsection{Study Setup}

\begin{figure}
    \centering
    \includegraphics[width=\linewidth]{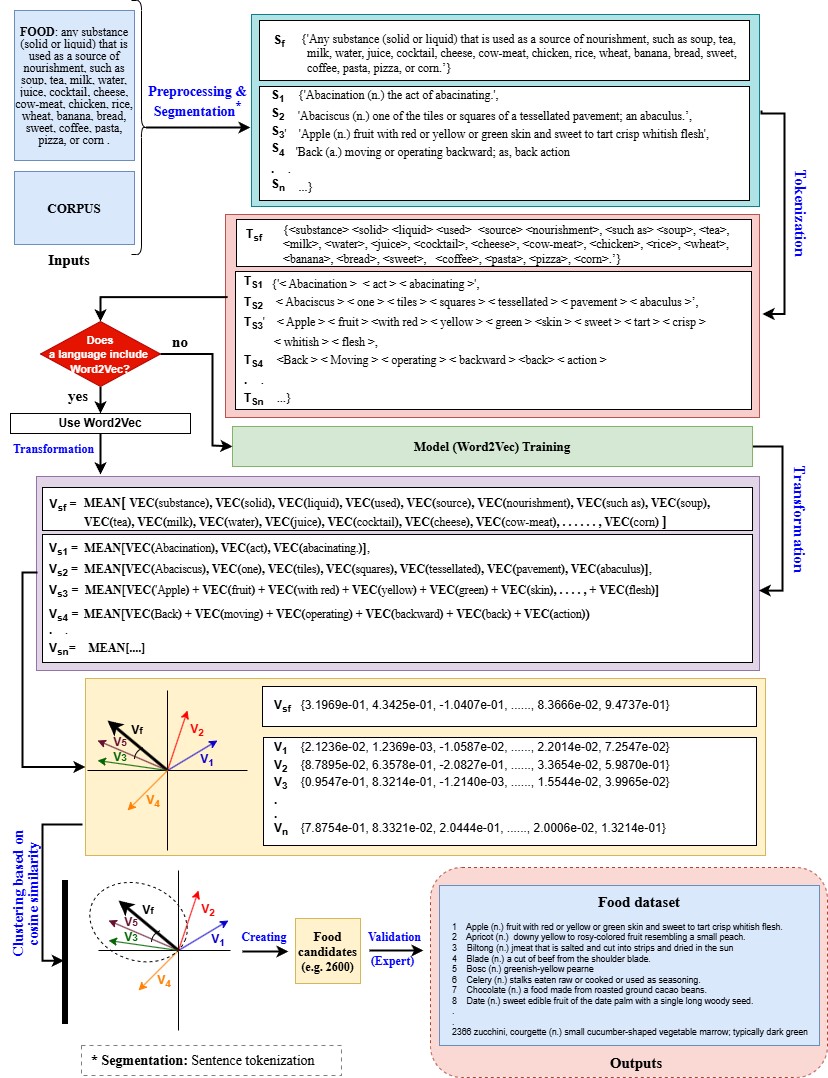}
    \caption{Flowchart of the semantic-field filtering method to collect food words from a corpus (e.g., Wikipedia).}
    \label{fig:BanjDsCreation}
\end{figure}

\noindent
\textbf{Task Generation} 

We developed two lexical datasets comprising 1,448~Indonesian terms and 812~Banjarese terms, both obtained using semantic filtering methods. For the Indonesian dataset, we employed a similar approach to that used for Arabic (\Cref{fig:ArabicDSCreation}). This method consists of three main steps. In \emph{Step 1}, the Kamus Bahasa Indonesia~\citep{kbbi} was used to extract all Indonesian terms. In \emph{Step 2}, the terms obtained from \emph{Step 1} were transformed into vector representations using IndoBERT~\citep{koto-etal-2020-indolem}. In \emph{Step 3}, these vectors were compared to the vector representation of the term ``food'' using cosine similarity to identify terms most similar to the centroid.
A custom Python script based on IndoBERT was developed to implement this methodology and collect food-related words in Indonesian from the Kamus Bahasa Indonesia dictionary. The script is available on GitHub\footnote{\url{https://github.com/HadiPTUK/developed_scripts/blob/main/indonesian_food_terms_script.py}}.


In the case of the Banjarese dataset, the absence of a high-quality dictionary or monolingual language model posed a significant challenge. To address this, we employed alternative resources—namely Word2Vec~\citep{mikolov2013efficientestimationwordrepresentations}, Wiktionary, and Wikipedia—to compile the dataset through a three-step process. \Cref{fig:BanjDsCreation} illustrates the methodological flowchart with examples, detailing the inputs, processing steps, and outputs. For clarity and ease of understanding, these examples are presented in \emph{English}.
In \emph{Step~1}, a Banjarese corpus was created by extracting data from the NLLB dataset~\citep{nllbteam2022languageleftbehindscaling} via NusaCrowd~\citep{cahyawijaya-etal-2023-nusacrowd}. This corpus was then used to build and train a static word embedding model using Word2Vec.
In \emph{Step~2}, a custom Python script\footnote{\url{https://github.com/HadiPTUK/developed_scripts/blob/main/banjarese_food_terms_script.py}}
 was employed to extract a list of terms and their definitions from both Wiktionary and Wikipedia. These terms were subsequently transformed into vector representations using the Word2Vec model developed in \emph{Step~1}.
In \emph{Step~3}, the vectorized terms were clustered around the term ``food'' and its definition, which served as the centroid. Cosine similarity was used to compute the distance between each term and the centroid, with a threshold set at 0.85. Terms exceeding this threshold were classified as food-related and added to the dataset.


\noindent
\textbf{Crowd selection.} 
Native speakers of Banjarese and Indonesian with sufficient bilingual proficiency were recruited as volunteers and screened using a two-phase quality control process: a proficiency test followed by the methodology illustrated in \Cref{fig2}. The final pool included six Indonesian native speakers fluent in Banjarese and six Banjarese native speakers proficient in Indonesian. All participants were university students. Each group was divided into two teams (G1 and G2) of three members.

\noindent
\textbf{Contribution collection and quality control.}
As in the English-Arabic experiment, we used the LingoGap platform and followed the guidelines outlined in \Cref{tab:guidelines_english_arabic} to enable qualified crowd workers to match Indonesian and Banjarese terms through micro-tasks. In both directions (\textit{Indonesian → Banjarese} and \textit{Banjarese → Indonesian}), participants were assigned 45~terms and 4~ACQs per 90-minute task. This configuration was based on our pilot experiments for this study.

In the \textit{Indonesian → Banjarese} experiment, 1,448~Indonesian terms were randomly distributed across 30~crowdsourcing tasks, as presented in \Cref{table_banjarese}, ensuring that no group focused exclusively on specific food categories. Each of Groups G1 and G2 completed 15~tasks. In the \textit{Banjarese → Indonesian} experiment, 330 Banjarese terms—non-overlapping with their Indonesian equivalents—were similarly distributed across 8~tasks, as shown in \Cref{table_indonesian}.

\noindent
\textbf{Task Validation.} To ensure the reliability and quality of volunteer responses in both translation directions, we employed the validation methodology illustrated in Figure~\ref{fig3}. A minimum agreement threshold of 0.7 was established and consistently exceeded by both participant groups. In the \textit{Banjarese~→~Indonesian} direction, certain entries lacking full (100\%) inter-annotator agreement were submitted to a linguistic expert for adjudication. For instance, the Banjarese word \textit{rabuk}—meaning ``\textit{a type of ground meat}''—was identified as a lexical gap by some participants; the expert corrected this by providing the Indonesian equivalent \textit{abon}. For further details, the corresponding datasets are available in the DataScientia repository: the Banjarese dataset\footnote{\url{https://ds.datascientia.eu/community/public/projects/c3d242b7-4ddc-419f-bffd-38d9f9760a05}} and the Indonesian dataset\footnote{\url{https://ds.datascientia.eu/community/public/projects/eacdb797-f8bf-45d4-8909-a4fd50ae8910}}.


%

\subsubsection{Study Results}
\label{IndonesianResults}
This section presents the datasets generated from two crowdsourcing experiments involving food-related terms in Indonesian and Banjarese.

In the \textit{Indonesian → Banjarese} experiment, crowd workers identified 750~lexical gaps, 507~equivalent terms, and 43~new Banjarese words not present in the original Banjarese input dataset. The annotation process demonstrated strong inter-annotator agreement, with an average Alpha score of 0.83. Additional details on the Banjarese crowdsourced data are provided in \Cref{table_banjarese}.

\begin{table}[h]
    \centering
    \caption{Crowdsourced Banjarese data summary.}
    \label{table_banjarese}
    \begin{tabular}{ccccccccc}
        \toprule
        \textbf{Task} & \multicolumn{2}{c}{\textbf{Gaps}} & \multicolumn{2}{c}{\textbf{Words}} & \multicolumn{2}{c}{\textbf{New Concepts}} & \multicolumn{2}{c}{\textbf{Alpha}} \\
        & G1 & G2 & G1 & G2 & G1 & G2 & G1 & G2 \\
        \midrule
        1 & 30 & 36 & 12 & 6 & 2 & 1 & 0.86 & 0.83 \\
        2 & 29 & 35 & 13 & 6 & 1 & 3 & 0.81 & 0.83 \\
        3 & 23 & 35 & 18 & 7 & 2 & 1 & 0.82 & 0.80 \\
        4 & 19 & 41 & 23 & 1 & 1 & 1 & 0.82 & 0.84 \\
        5 & 22 & 42 & 19 & 2 & 2 & 0 & 0.83 & 0.79 \\
        6 & 20 & 35 & 21 & 9 & 3 & 0 & 0.91 & 0.81 \\
        7 & 22 & 32 & 20 & 9 & 2 & 2 & 0.83 & 0.82 \\
        8 & 25 & 34 & 17 & 8 & 1 & 1 & 0.82 & 0.85 \\
        9 & 11 & 26 & 32 & 14 & 0 & 3 & 0.82 & 0.82 \\
        10 & 21 & 28 & 21 & 13 & 1 & 2 & 0.82 & 0.81 \\
        11 & 19 & 32 & 25 & 10 & 0 & 1 & 0.85 & 0.83 \\
        12 & 11 & 32 & 31 & 10 & 1 & 1 & 0.85 & 0.83 \\
        13 & 12 & 28 & 30 & 13 & 2 & 3 & 0.82 & 0.87 \\
        14 & 7 & 30 & 34 & 11 & 2 & 3 & 0.88 & 0.84 \\
        15 & 8 & 5 & 34 & 38 & 1 & 0 & 0.82 & 0.82 \\
        \midrule
        \textbf{Total} & 279 & 471 & 350 & 157 & 21 & 22 & & \\
        \textbf{Average} & & & & & & & 0.84 & 0.83 \\
        \bottomrule
    \end{tabular}
\end{table}

In the \textit{Banjarese → Indonesian} direction, the experiment revealed 201~lexical gaps, 98~equivalent terms, and 30~new Indonesian words that were absent from the original Indonesian dataset. This task also exhibited good annotator consistency, with an average Alpha score of 0.83. Further information on the Indonesian crowdsourced data is available in \Cref{table_indonesian}.

\begin{table}[h]
    \caption{Crowdsourced Indonesian data summary.}
    \label{table_indonesian}
    \centering
    \begin{tabular}{ccccccccc}
        \toprule
        \textbf{Task} & \multicolumn{2}{c}{\textbf{Gaps}} & \multicolumn{2}{c}{\textbf{Words}} & \multicolumn{2}{c}{\textbf{Concepts}} & \multicolumn{2}{c}{\textbf{Alpha}} \\
        & G1 & G2 & G1 & G2 & G1 & G2 & G1 & G2 \\
        \midrule
        1 & 26 & 25 & 15 & 17 & 4 & 3 & 0.82 & 0.87 \\
        2 & 28 & 29 & 14 & 12 & 3 & 4 & 0.85 & 0.81 \\
        3 & 28 & 28 & 12 & 13 & 5 & 4 & 0.82 & 0.84 \\
        4 & 30 & 7  & 11 & 4  & 4 & 3 & 0.81 & 0.81 \\
        \midrule
        \textbf{Total} & 112 & 89 & 52 & 46 & 16 & 14 & & \\
        \textbf{Avg}   &     &     &     &     &     &     & 0.83 & 0.83 \\
        \bottomrule
    \end{tabular}
\end{table}

\subsubsection{Lexical Diversity Evaluation: Overlap-Based Metric}
This section examines the overlap between Indonesian and Banjarese in the food domain. As shown in \Cref{fig:venn_idn_bjn}, the intersection of the Indonesian and Banjarese languages for a shared coverage of 40.9\%. The number of lexicalisations in Indonesian is 1,478~(1,448~words in the original Indonesian input dataset and 30~new words), and in Banjarese is 855~(812~words in the original Banjarese input dataset and 43~new words). Also, 605~of these lexical units~(507~words were explored in the experiment of \textit{Indonesian → Banjarese}, 98~words were identified in the experiment of \textit{Banjarese → Indonesian}) are included in both languages. For example, \Cref{eq:crowdsourcing_overlap} calculates the overlap between Indonesian and Banjarese (both represented in ISO 639-3 as ``ind'' and ``bjn'', respectively) in the food domain ($F$) as follows:

\newpage

\begin{figure}[h]
    \centering
    \includegraphics[width=0.5\linewidth]{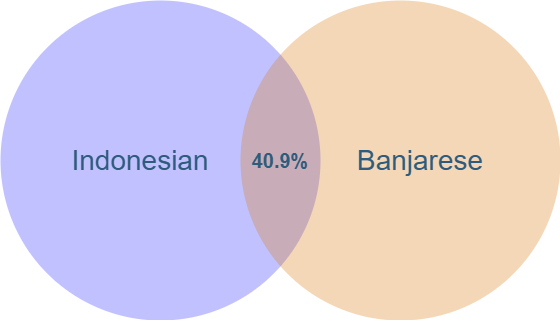}
    \caption{The overlap (percentage of shared lexicalizations) for Indonesian and Banjarese languages.}
    \label{fig:venn_idn_bjn}
\end{figure}

\begin{equation}
    \mathrm{overlap}(F, \mathrm{ind}, \mathrm{bjn})=\frac{|\mathrm{LexCons}(F, \mathrm{ind})\cap \mathrm{LexCons}(F, \mathrm{bjn})|}{\textrm{max}(|\mathrm{LexCons}(F, \mathrm{ind})|, |\mathrm{LexCons}(F, \mathrm{bjn})|)}
\end{equation}

\begin{equation}
    \mathrm{overlap}(F, \mathrm{ind}, \mathrm{bjn})=\frac{605}{\mathrm{max}(1478, 855)}=\frac{605}{1478}=40.9\%
\end{equation}

While both Indonesian and Banjarese are part of the Austronesian language family, they have different linguistic histories and influences. Banjarese has absorbed vocabulary from Dayak languages, Malay dialects, and other indigenous languages of Kalimantan, while Indonesian, as the national language, has been influenced by Malay, Javanese, Dutch, Arabic, and other foreign languages. 
Furthermore, these languages exist on different islands in Indonesia; Banjarese is located on the southern part of Borneo Island, and the Indonesian language is spoken on Sumatra Island \citep{sneddon2003}, so this geographical barrier restricts interactions between speakers, and each language has developed within its own speech community. These historical and geographical influences leads to differences in vocabulary, including food terms.

\subsection{Quality of Crowdsourced Data: A Comparison with LLMs Annotation}
\label{SecLLMs1}
To complement the crowdsourced annotations, we evaluated whether LLMs could perform the same lexical gap identification task, aiming to assess how effectively current LLMs capture cultural and semantic nuances typically recognized by native speakers. This section explores the use of LLMs as annotation tools in linguistic research, focusing on the construction of small, diversity-aware datasets that serve as benchmarks for evaluating the reliability of diversity-related data obtained through crowdsourcing. Specifically, we assess the translatability of lexical items to examine lexical diversity and semantic equivalence across source and target languages, using LLMs to capture nuanced cross-linguistic correspondences.



Recent studies underscore the growing effectiveness of LLMs, particularly GPT-based models, in text annotation tasks. For instance, GPT-3 and GPT-4 have been employed to annotate political tweets and identify propaganda \citep{ding2022, hasanain2023}, while Google’s Gemini Pro has been used for structured data extraction in scientific texts \citep{sayeed2024annotating}. Comparative evaluations indicate that GPT models frequently outperform both crowdsourced annotators and domain experts in tasks such as political stance detection \citep{alizadeh2023, tornberg2023}. Building on this body of work, we utilize GPT-4, DeepSeek, and Gemini to annotate cross-linguistic data, with a focus on lexical diversity and semantic equivalence. GPT-4 is prioritized due to its demonstrated superior performance.

\begin{figure}[htbp]
    \centering
    \includegraphics[width=0.7\linewidth]{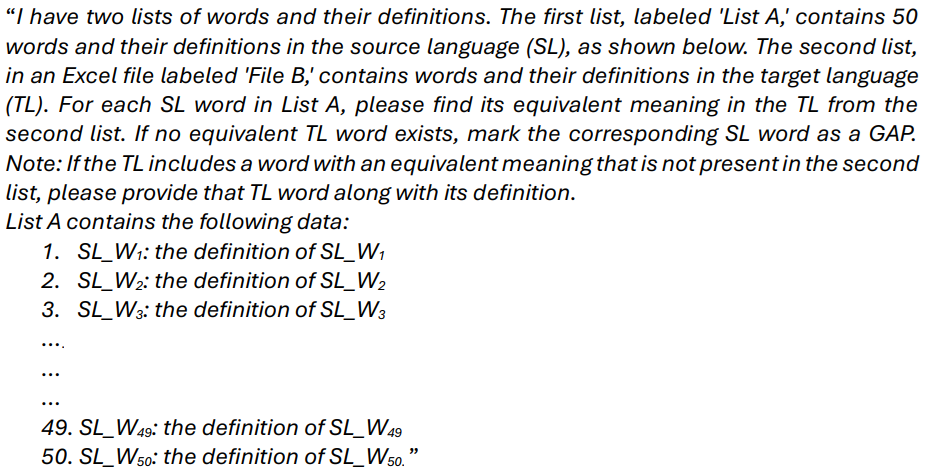}
    \caption{The template used to prompt each of the three LLMs—GPT-4o, DeepSeek-V3, and Gemini 2.0 Flash.}
    \label{fig:GPTprompt}
\end{figure}

Our annotation process adapts the crowdsourcing methodology described in \Cref{secMethod}, incorporating two key modifications. \textit{First}, we replace crowdsourcing with an \emph{Annotation Phase}, during which LLMs perform the annotation. Each task consists of (50) \footnote{We conducted three experiments under different prompt configurations: \textit{Experiment~1}: Each prompt contained a single SL word. \textit{Experiment~2}: Each prompt included 10 SL words, following the same procedure. \textit{Experiment~3}: Each prompt contained 50 SL words, applying the same method. Across all experiments, annotation accuracy for SL words remained consistent. For instance, in all cases in \Cref{EnglishArabicLLs}, the SL term \AR{مهلبية} meaning ``\textit{a dessert made from milk, starch, and sugar}'' was matched to the TL term \textit{pudding}, rather than being classified as a lexical gap.} SL lexical items within the specified SF, paired with TL candidates. The objective is to determine whether a TL equivalent exists. If it does, the TL entry is selected or proposed as a new term; if not, the SL item is marked as a lexical gap. \Cref{fig:GPTprompt} presents the prompt template used. \textit{Second}, we rely exclusively on expert validation to assess the accuracy of identified equivalents and lexical gaps. This method is applied in two food-domain case studies: \emph{English–Arabic} and \emph{Indonesian–Banjarese}. These language pairs were also examined using our crowdsourcing approach, as detailed in \cref{secArabicExp} and \cref{secIndonesianExp}.


Each experiment was conducted using GPT-4o, DeepSeek-V3, and Gemini 2.0 Flash via their web interfaces. The models were neither fine-tuned nor externally assisted, ensuring an authentic comparison with human annotators. Each query followed the same structure (as shown in \Cref{fig:GPTprompt}) and input format as the human task to maintain methodological consistency. Specifically, each LLM received the SL word and its definition (gloss) as input and was prompted to determine whether an equivalent existed in the TL.


As demonstrated in the two case studies, the results indicate that LLMs struggle to identify lexical gaps, particularly in low-resource and culturally diverse contexts. Three key findings emerged:

\begin{itemize}
\item \textit{Hallucination of TL terms}: LLMs occasionally generated plausible but nonexistent TL words when no equivalent existed, suggesting overgeneralization.
\item \textit{Failure in culturally grounded semantics}: The models frequently misinterpreted culturally embedded terms (e.g., food items), generating translations that were overly generic or semantically inaccurate.
\item \textit{Lack of sociocultural awareness}: Unlike human annotators, LLMs were unable to infer when the absence of a TL equivalent reflected a deeper cultural or linguistic divergence.
\end{itemize}

While the LLM performed well on straightforward lexical mappings, it demonstrated limited sensitivity to subtle semantic distinctions and sociocultural contexts. These findings reinforce the paper’s central argument that human-in-the-loop crowdsourcing remains essential for capturing linguistic diversity and validating lexical gaps. Although LLMs can assist with data generation and pre-filtering, they cannot yet replicate the nuanced reasoning of native speakers in cross-linguistic annotation tasks.

\subsubsection{Case Study: Food Terminology in English and Arabic}
\label{EnglishArabicLLs}
This case study investigates cross-linguistic lexical diversity between English and Arabic through two experiments. The first experiment (\textit{English → Arabic}) examines how English food terms are lexicalized in Arabic. Fifty culturally specific English food terms were selected from a dataset of 2,364~terms compiled in \Cref{secArabicExp}. These terms were compared against an Arabic dataset of 1,607~words (also created in \Cref{secArabicExp}) using GPT-4o, DeepSeek-V3, and Gemini 2.0 Flash. Each model performed zero-shot annotations, either identifying an Arabic equivalent or marking the term as a lexical gap. 


The second experiment (\emph{Arabic → English}) reversed the direction of analysis. Fifty Arabic-specific food terms were selected and matched against the English dataset using the same language models. Each model annotated the Arabic terms by either identifying English equivalents or indicating lexical gaps.

Expert validation was conducted to ensure annotation accuracy. An assistant professor specializing in lexical semantics and a native speaker of Arabic validated the Arabic equivalents and lexical gaps. An Arabic-speaking university lecturer based in the United Kingdom reviewed the English annotations. \Cref{tab:accuracy_gpt4} presents the models’ annotation \emph{accuracy}, defined as the number of correctly identified equivalents or gaps divided by the total number of source terms. Two common error types were observed: (1) providing \textit{incorrect TL equivalents}, and (2) offering \textit{literal translations} instead of correctly identifying lexical gaps.

For instance, in the \emph{Arabic → English} experiment, the Arabic term \AR{الأبيضان} ``\textit{water and yogurt}'' was mistranslated by GPT-4o as ``\textit{the two whites}''.” Similarly, \AR{كيبة} ``\textit{a dish made of ground meat and rice or wheat}'' was translated literally as \textit{kibbeh} by all models, although the term lacks recognition in standard English lexicons.

In the \emph{English → Arabic} experiment, \textit{malt liquor}, meaning ``\textit{a strong lager}'', was translated by GPT-4o and DeepSeek-V3 as \AR{مشروب شعير} ``\textit{barley drink}'', omitting the alcoholic context. Likewise, the term \textit{stout}, meaning ``\textit{a dark ale}'', was transliterated as \AR{ستاوت} by Gemini 2.0 Flash and DeepSeek-V3—a term not attested in Arabic lexical resources.

\begin{table}[h]
    \centering
    \small
    \begin{tabular}{cccccc}
        \toprule
        \textbf{Exp. No} & \textbf{Source Language} & \textbf{Target Language} & \textbf{GPT-4o} & \textbf{DeepSeek-V3} & \textbf{Gemini 2.0 Flash} \\
        \midrule
        1 & English & Arabic & 18 & 38 & 14 \\
        2 & Arabic & English & 42 & 46 & 38 \\
        3 & Indonesian & Banjarese & 8 & 10 & 36 \\
        4 & Banjarese & Indonesian & 40 & 46 & 58 \\
        \bottomrule
    \end{tabular}
    \caption{Percentage accuracy of validated data collected by GPT-4o, DeepSeek-V3, and Gemini 2.0 Flash models.}
    \label{tab:accuracy_gpt4}
\end{table}


\subsubsection{Case Study: Food Terminology in Indonesian and Banjarese}
\label{SecIndoBanjLLs}
This case study investigates lexical diversity in the food domain between Indonesian and Banjarese through two experiments. In the \textit{Indonesian → Banjarese} experiment, we selected 50 culturally specific Indonesian food terms from a dataset of 1,448 entries and compared them against a Banjarese dataset comprising 812 terms. Both datasets were constructed as described in Section \ref{secIndonesianExp}. Using zero-shot prompts, GPT-4o, DeepSeek-V3, and Gemini 2.0 Flash annotated each term by either identifying a Banjarese equivalent or indicating a lexical gap.

The \textit{Banjarese → Indonesian} experiment mirrored this approach in reverse, selecting 50 Banjarese-specific terms and comparing them against the Indonesian dataset. Each language model annotated the terms with corresponding Indonesian equivalents or flagged lexical gaps.

Annotations were reviewed by two native speakers: a university linguistics instructor validated the Banjarese results, while an Indonesian master’s student specializing in lexical semantics validated the Indonesian results. Accuracy scores for each model in both translation directions are reported in \Cref{tab:accuracy_gpt4}, based on expert validation. As in the English–Arabic study, the models frequently made two types of errors: (1) selecting incorrect TL equivalents, and (2) providing literal translations where lexical gaps should have been identified. Examples of such errors include:

\begin{itemize}
    \item GPT-4o mistranslated the Indonesian word \textit{Beras} ``\textit{uncooked rice}'' as \textit{Karak} ``\textit{hardened rice}''.
    \item  DeepSeek-V3 rendered \textit{Papaya} ``\textit{a large oval tropical fruit with yellowish flesh}'' as \textit{Gandis} ``\textit{yellow mangosteen}''.
    \item Gemini 2.0 Flash mistranslated the Banjarese word \textit{Balinjan} ``\textit{Tomato}'' as \textit{Terung} ``\textit{Eggplant}''.
     \item GPT-4o  misinterpreted the Banjarese term \textit{Janar} ``\textit{Turmeric}'' as \textit{Jahe} ``\textit{Ginger}''.
\end{itemize}

\begin{figure}[h]
    \centering
    \includegraphics[width=0.7\columnwidth]{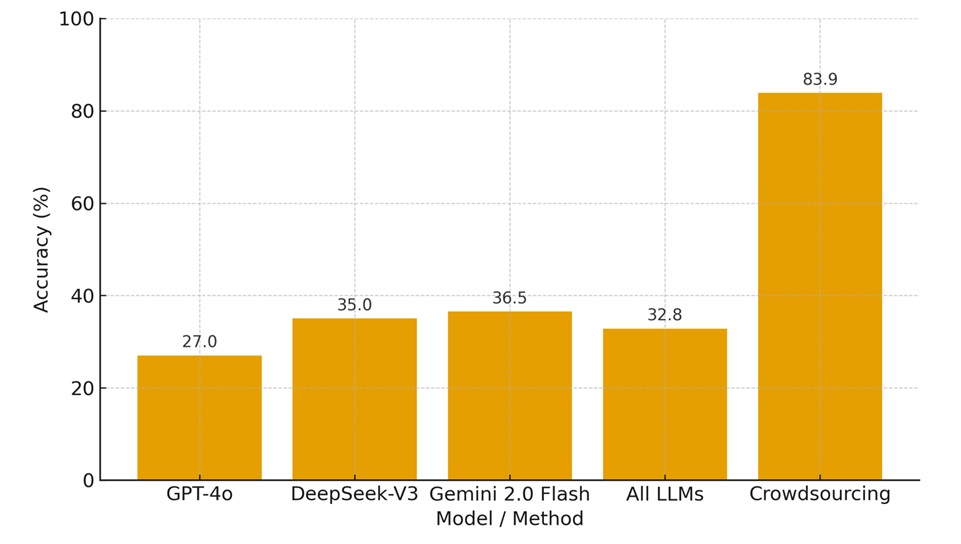} 
    \caption{Accuracy comparison between LLM and crowdsourced annotations across collected lexical gaps and words}
    \label{fig:bar_LLMs_crowdsourcing}
\end{figure}

These findings, together with those from the English–Arabic experiment, underscore the challenges that LLMs face in accurately capturing nuanced or culturally specific lexical meanings—particularly across low-resource language pairs. The bar chart in \Cref{fig:bar_LLMs_crowdsourcing} illustrates the average accuracy across four lexicalisation experiments for three LLMs (GPT-4o, DeepSeek-V3, and Gemini 2.0 Flash), their combined mean (“All LLMs”), and human crowdsourcing. While the LLMs achieve moderate accuracy ranging from 27\% to 37\%, their combined average of approximately 33\% remains substantially lower than the 83.9\% accuracy achieved through crowdsourcing. This comparison clearly reveals a pronounced performance gap, indicating that human participants consistently outperform current LLMs in identifying and translating culturally specific lexical items.

\section{Perspectives on Crowdsourcing}
\label{secPersCrowd}
Our crowdsourcing approach offers significant advantages for investigating linguistic diversity. It enables the \textit{systematic} identification of \textit{lexical gaps}, \textit{culture-specific concepts}, and \textit{equivalent word meanings}, thereby providing valuable insights into language-specific phenomena. The involvement of native speakers ensures contextual precision and captures subtle linguistic and cultural nuances.

This method supports the creation of datasets that are sensitive to lexical diversity and facilitates \textit{bidirectional} exploration between source and target languages, particularly in socially relevant domains such as food. It is both \textit{scalable} and \textit{adaptable}, capable of handling datasets of varying sizes by dividing them into manageable micro-tasks. This task design enhances efficiency and accessibility, even for \textit{non-expert contributors}. For example, in the English-Arabic experiment, each task comprised 35~SL words, while the Indonesian-Banjarese task included 45. Quality control was maintained through real-time validation, \textit{time tracking}, and two \textit{Krippendorff’s Alpha–based strategies}: a filtering algorithm during pilot testing to exclude low-quality contributors, and cross-validation of responses using IAA scores across contributor groups.

The \textit{LingoGap} platform streamlines this process with a user-friendly interface, clear instructions, and well-structured workflows. Features such as response time tracking and detailed task logs enhance \textit{transparency} and improve data reliability.


In the long term, the proposed crowdsourcing methodology provides a scalable and sustainable pathway toward more inclusive NLP, enabling the integration of culturally grounded lexical data from underrepresented languages. This integration contributes to richer \textit{multilingual embeddings}, improved \textit{cross-lingual transfer} in large language models, and the sustainable development of linguistic resources across both high- and low-resource languages.

Despite these strengths, the approach presents several challenges. \textit{Timing} and contributor availability significantly affect progress, especially during exam periods, holidays, or personal events. For instance, one task (Task~10 by $G_4$ in the Arabic-to-English experiment) was delayed by two weeks due to a contributor's maternity leave, while another (Task~7 by $G_1$ in the Indonesian-to-Banjarese experiment) was postponed because of a family event. These real-world disruptions underscore the importance of flexible task management and careful scheduling, reaffirming earlier findings that the timing of a crowdsourcing campaign is critical, as contributors' availability can be influenced by various external factors \citep{christoforou2021s}.

Recruiting a sufficiently diverse pool of participants remains a challenge, impacting the robustness of evaluation. Additional difficulties arise from the \textit{ambiguity} of certain SL terms—particularly those that fall between well-defined concepts and lexical gaps. For example, the Arabic word \AR{مهلبية} ``\textit{a type of dessert made from milk, starch, and sugar}'' and the Banjarese word ``rabuk'' ``\textit{ground meat}'' posed classification challenges.

\textit{Newly coined} and \textit{rare terms} further complicate validation process, as they are often under-documented. The approach also struggles to capture \textit{connotative meanings} and affective or metaphorical nuances, which are highly context-dependent and shaped by individual sociocultural backgrounds—dimensions not easily evaluated using standard IAA metrics \cite{levinson2000presumptive,wierzbicka1996semantics}.

Lastly, \textit{sociolinguistic variation}—including differences based on region, class, age, or gender—is often underrepresented due to demographic biases inherent in many crowdsourcing platforms \cite{eckert2000linguistic}. Future research should consider employing stratified sampling and sociolinguistically sensitive task design to enhance representativeness.

Together, these perspectives highlight both the potential and the limitations of crowdsourcing in linguistic research. With thoughtful design and continuous refinement, crowdsourcing can serve as a powerful tool for documenting and analyzing lexical diversity across languages.

\section{Conclusion}
\label{secConclusion}

This study addresses the ongoing challenge of linguistic diversity and underrepresentation in multilingual lexical-semantic resources (LSRs), which are essential for many NLP tasks. Current LSRs often reflect English-centric biases and fail to adequately represent lexical gaps, especially in low-resource languages.

To overcome these issues, we introduced a scalable crowdsourcing methodology to systematically collect data on lexical equivalence and gaps across languages. Our approach consists of three key steps: (1)~generating input datasets using linguistic resources or language models, (2)~gathering responses through the LingoGap crowdsourcing platform, and (3)~validating results via IAA and expert review. LingoGap supports the entire process, from task creation to contributor engagement and quality assurance.

We demonstrated the effectiveness of this method through two \textit{large-scale case studies} on food terminology: English–Arabic and Indonesian–Banjarese. These case studies highlight the approach’s scalability and its ability to capture lexical diversity in both high- and low-resource languages. Overall, we identified 3,091~lexical gaps and collected 1,735~words across the four languages.

Our findings contribute to building more inclusive and accurate multilingual LSRs, with implications for NLP applications including machine translation, word sense disambiguation, and other NLP tasks.
 

Future work should extend this methodology to a \textit{broader range of languages and dialects}, particularly underrepresented and typologically diverse ones, to further evaluate its scalability and adaptability. Additionally, exploring \textit{culturally rich semantic domains}—such as body parts \citep{wierzbicka2007}, colors \citep{roberson2005}, visual concepts \citep{giunchiglia2021}, emotions, and social values—can provide deeper insights into cross-linguistic and cross-cultural patterns.
 

Enhancing the data collection process through \textit{real-time expert validation} and incorporating features such as \textit{gamification} and \textit{mobile accessibility} can further improve data quality and increase participation—particularly by engaging broader demographics, including students and the general public.

Another promising direction involves integrating diversity-aware lexical data into LLMs and other NLP systems during both training and evaluation to enhance their cultural and linguistic sensitivity. Such enriched resources can also support a wide range of cross-lingual NLP tasks—including more equitable machine translation, cross-lingual information retrieval, and semantic parsing—as well as interdisciplinary research in education, sociolinguistics, cultural studies \citep{ono2023consideration}, and linguistic anthropology, particularly for low-resource languages.


In summary, this work offers a practical, scalable framework for advancing linguistic inclusivity in multilingual NLP, promoting more equitable access to language technologies for diverse language communities.

\section*{Conflict of Interest Statement}
The authors declare that the research was conducted in the absence of any commercial or financial relationships that could be construed as a potential conflict of interest.


\section*{Author Contributions}
FG, HK, and JO conceptualized and supervised the study. FG provided overall supervision. HK wrote the original manuscript draft and conducted the English-Arabic experiment. HK also prepared and formatted the English and Arabic food datasets and validated the collected data on lexical diversity in Arabic. SD conducted the Indonesian-Banjarese experiment and validated the results. FG, GB, HK, and SD analyzed the data collected across the four languages. GB validated the identified new concepts FG, HK, GB, JO, and SD reviewed and revised the manuscript. All authors contributed to the research and approved the final submitted version.




\section*{Acknowledgments}
We would like to thank the University of Trento and Palestine Technical University—Kadoori for their support.  
We are also grateful to Prof. Rusma Noortyani from Lambung Mangkurat University, Indonesia, for her assistance in recruiting students for the Indonesian–Banjarese experiment.


\section*{Data Availability Statement}
The diversity-aware food datasets generated and analyzed in this study are publicly available in the DataScientia repository at the following links: 

\begin{itemize}
    \item English-to-Arabic dataset: \textit{\url{https://ds.datascientia.eu/community/public/projects/aef86b5b-5742-4529-9c46-71b5142f91bd}}
    \item Arabic-to-English dataset: \textit{\url{https://ds.datascientia.eu/community/public/projects/db9ec453-4775-43d6-a89a-85a9600b9781}}
    \item Indonesian-to-Banjarese dataset: \textit{\url{https://ds.datascientia.eu/community/public/projects/c3d242b7-4ddc-419f-bffd-38d9f9760a05}}
    \item Banjarese-to-Indonesian dataset: \textit{\url{https://ds.datascientia.eu/community/public/projects/eacdb797-f8bf-45d4-8909-a4fd50ae8910}}
\end{itemize}

\bibliographystyle{apalike}
\bibliography{main}  

\begin{thebibliography}{}

\bibitem[Albala, 2011]{albala2011}
Albala, K. (2011).
\newblock {\em Food Cultures of the World Encyclopedia:[4 Volumes]}.
\newblock Bloomsbury Publishing USA.

\bibitem[Alizadeh et~al., 2023]{alizadeh2023}
Alizadeh, M., Kubli, M., Samei, Z., Dehghani, S., Bermeo, J.~D., Korobeynikova, M., and Gilardi, F. (2023).
\newblock Open-source large language models outperform crowd workers and approach chatgpt in text-annotation tasks.
\newblock {\em arXiv preprint arXiv:2307.02179}, 101.

\bibitem[Armanios and Ergene, 2018]{armanios2018}
Armanios, F. and Ergene, B. (2018).
\newblock {\em Halal food: A history}.
\newblock Oxford University Press.

\bibitem[Artstein and Poesio, 2008]{artstein2008}
Artstein, R. and Poesio, M. (2008).
\newblock Inter-coder agreement for computational linguistics.
\newblock {\em Computational linguistics}, 34(4):555--596.

\bibitem[Ashley et~al., 2004]{ashley2004}
Ashley, B., Hollows, J., Jones, S., and Taylor, B. (2004).
\newblock {\em Food and cultural studies}.
\newblock Routledge.

\bibitem[Barbouch et~al., 2021]{barbouch2021}
Barbouch, M., Verberne, S., and Verhoef, T. (2021).
\newblock Wn-bert: Integrating wordnet and bert for lexical semantics in natural language understanding.
\newblock {\em Computational Linguistics in the Netherlands Journal}, 11:105--124.

\bibitem[Bella et~al., 2022]{bella2022}
Bella, G., Batsuren, K., Khishigsuren, T., and Giunchiglia, F. (2022).
\newblock Linguistic diversity and bias in online dictionaries.
\newblock In Lena, K., editor, {\em Frontiers in African Digital Research}, pages 173--186. Institute of African Studies.

\bibitem[Bella et~al., 2024]{bella2024tackling}
Bella, G., Helm, P., Koch, G., and Giunchiglia, F. (2024).
\newblock Tackling language modelling bias in support of linguistic diversity.
\newblock In {\em Proceedings of the 2024 ACM Conference on Fairness, Accountability, and Transparency}, FAccT '24, page 562–572, New York, NY, USA. Association for Computing Machinery.

\bibitem[Benjamin and Radetzky, 2014]{benjamin2014}
Benjamin, M. and Radetzky, P. (2014).
\newblock Multilingual lexicography with a focus on less-resourced languages: Data mining, expert input, crowdsourcing, and gamification.
\newblock In {\em 9th edition of the Language Resources and Evaluation Conference}.

\bibitem[Biemann and Nygaard, 2010]{biemann2010}
Biemann, C. and Nygaard, V. (2010).
\newblock Crowdsourcing wordnet.
\newblock In {\em The 5th International Conference of the Global WordNet Association (GWC-2010)}.

\bibitem[Bond and Foster, 2013]{bond2013}
Bond, F. and Foster, R. (2013).
\newblock Linking and extending an open multilingual {W}ordnet.
\newblock In Schuetze, H., Fung, P., and Poesio, M., editors, {\em Proceedings of the 51st Annual Meeting of the Association for Computational Linguistics (Volume 1: Long Papers)}, pages 1352--1362, Sofia, Bulgaria. Association for Computational Linguistics.

\bibitem[Brualdi, 2004]{brualdi2004}
Brualdi, R.~A. (2004).
\newblock {\em Introductory combinatorics}.
\newblock Pearson Education India.

\bibitem[Cahyawijaya et~al., 2023]{cahyawijaya-etal-2023-nusacrowd}
Cahyawijaya, S., Lovenia, H., Aji, A.~F., Winata, G., Wilie, B., Koto, F., Mahendra, R., Wibisono, C., Romadhony, A., Vincentio, K., Santoso, J., Moeljadi, D., Wirawan, C., Hudi, F., Wicaksono, M.~S., Parmonangan, I., Alfina, I., Putra, I.~F., Rahmadani, S., Oenang, Y., Septiandri, A., Jaya, J., Dhole, K., Suryani, A., Putri, R.~A., Su, D., Stevens, K., Nityasya, M.~N., Adilazuarda, M., Hadiwijaya, R., Diandaru, R., Yu, T., Ghifari, V., Dai, W., Xu, Y., Damapuspita, D., Wibowo, H., Tho, C., Karo~Karo, I., Fatyanosa, T., Ji, Z., Neubig, G., Baldwin, T., Ruder, S., Fung, P., Sujaini, H., Sakti, S., and Purwarianti, A. (2023).
\newblock {N}usa{C}rowd: Open source initiative for {I}ndonesian {NLP} resources.
\newblock In Rogers, A., Boyd-Graber, J., and Okazaki, N., editors, {\em Findings of the Association for Computational Linguistics: ACL 2023}, pages 13745--13818, Toronto, Canada. Association for Computational Linguistics.

\bibitem[Catford, 1965]{catford1965linguistic}
Catford, J.~C. (1965).
\newblock A linguistic theory of translation.

\bibitem[Christoforou et~al., 2021]{christoforou2021s}
Christoforou, E., Barlas, P., and Otterbacher, J. (2021).
\newblock It’s about time: A view of crowdsourced data before and during the pandemic.
\newblock In {\em Proceedings of the 2021 CHI Conference on Human Factors in Computing Systems}, CHI '21, New York, NY, USA. Association for Computing Machinery.

\bibitem[{\v{C}}ibej and Arhar~Holdt, 2019]{cibej2019}
{\v{C}}ibej, J. and Arhar~Holdt, {\v{S}}. (2019).
\newblock Repel the syntruders! a crowdsourcing cleanup of the thesaurus of modern slovene.
\newblock In {\em Proceedings of the ELex 2019 Conference: Electronic lexicography in the 21st century, Sintra, Portugal}.

\bibitem[Ding et~al., 2023]{ding2022}
Ding, B., Qin, C., Liu, L., Chia, Y.~K., Li, B., Joty, S., and Bing, L. (2023).
\newblock Is {GPT}-3 a good data annotator?
\newblock In Rogers, A., Boyd-Graber, J., and Okazaki, N., editors, {\em Proceedings of the 61st Annual Meeting of the Association for Computational Linguistics (Volume 1: Long Papers)}, pages 11173--11195, Toronto, Canada. Association for Computational Linguistics.

\bibitem[Dryer and Haspelmath, 2013]{dryer2013}
Dryer, M.~S. and Haspelmath, M., editors (2013).
\newblock {\em WALS Online (v2020.3)}.
\newblock Zenodo.

\bibitem[Eckert, 2000]{eckert2000linguistic}
Eckert, P. (2000).
\newblock {\em Linguistic Variation as Social Practice: The Linguistic Construction of Identity in Belten High}.
\newblock Blackwell Publishers, Malden, MA.

\bibitem[El-Haj et~al., 2015]{elhaj2015}
El-Haj, M., Kruschwitz, U., and Fox, C. (2015).
\newblock Creating language resources for under-resourced languages: methodologies, and experiments with arabic.
\newblock {\em Language Resources and Evaluation}, 49:549--580.

\bibitem[Fellbaum and Vossen, 2012]{fellbaum2012}
Fellbaum, C. and Vossen, P. (2012).
\newblock Challenges for a multilingual wordnet.
\newblock {\em Language Resources and Evaluation}, 46:313--326.

\bibitem[Fi{\v{s}}er et~al., 2014]{fivser2014}
Fi{\v{s}}er, D., Tav{\v{c}}ar, A., and Erjavec, T. (2014).
\newblock slo{WC}rowd: A crowdsourcing tool for lexicographic tasks.
\newblock In Calzolari, N., Choukri, K., Declerck, T., Loftsson, H., Maegaard, B., Mariani, J., Moreno, A., Odijk, J., and Piperidis, S., editors, {\em Proceedings of the Ninth International Conference on Language Resources and Evaluation ({LREC}'14)}, pages 3471--3475, Reykjavik, Iceland. European Language Resources Association (ELRA).

\bibitem[Franklin et~al., 2011]{franklin2011}
Franklin, M.~J., Kossmann, D., Kraska, T., Ramesh, S., and Xin, R. (2011).
\newblock Crowddb: answering queries with crowdsourcing.
\newblock In {\em Proceedings of the 2011 ACM SIGMOD International Conference on Management of Data}, SIGMOD '11, page 61–72, New York, NY, USA. Association for Computing Machinery.

\bibitem[Freihat et~al., 2024]{freihat2024}
Freihat, A.~A., Khalilia, H., Bella, G., and Giunchiglia, F. (2024).
\newblock Advancing the {A}rabic {W}ord{N}et: Elevating content quality.
\newblock In Al-Khalifa, H., Darwish, K., Mubarak, H., Ali, M., and Elsayed, T., editors, {\em Proceedings of the 6th Workshop on Open-Source Arabic Corpora and Processing Tools (OSACT) with Shared Tasks on Arabic LLMs Hallucination and Dialect to MSA Machine Translation @ LREC-COLING 2024}, pages 74--83, Torino, Italia. ELRA and ICCL.

\bibitem[Ganbold et~al., 2018]{ganbold2018}
Ganbold, A., Chagnaa, A., and Bella, G. (2018).
\newblock Using crowd agreement for {W}ordnet localization.
\newblock In Calzolari, N., Choukri, K., Cieri, C., Declerck, T., Goggi, S., Hasida, K., Isahara, H., Maegaard, B., Mariani, J., Mazo, H., Moreno, A., Odijk, J., Piperidis, S., and Tokunaga, T., editors, {\em Proceedings of the Eleventh International Conference on Language Resources and Evaluation ({LREC} 2018)}, Miyazaki, Japan. European Language Resources Association (ELRA).

\bibitem[Gantar and Krek, 2011]{gantar2011}
Gantar, P. and Krek, S. (2011).
\newblock Slovene lexical database.
\newblock {\em Natural language processing, multilinguality}, pages 72--80.

\bibitem[Georgakopoulos et~al., 2022]{georgakopoulos2022}
Georgakopoulos, T., Grossman, E., Nikolaev, D., and Polis, S. (2022).
\newblock Universal and macro-areal patterns in the lexicon: A case-study in the perception-cognition domain.
\newblock {\em Linguistic Typology}, 26(2):439--487.

\bibitem[Giunchiglia and Bagchi, 2021]{giunchiglia2021}
Giunchiglia, F. and Bagchi, M. (2021).
\newblock Classifying concepts via visual properties.

\bibitem[Giunchiglia et~al., 2018]{giunchiglia2018one}
Giunchiglia, F., Batsuren, K., and Alhakim~Freihat, A. (2018).
\newblock One world-seven thousand languages (best paper award, third place).
\newblock In {\em International Conference on Computational Linguistics and Intelligent Text Processing}, pages 220--235. Springer.

\bibitem[Giunchiglia et~al., 2017]{giunchiglia2017understanding}
Giunchiglia, F., Batsuren, K., Bella, G., et~al. (2017).
\newblock Understanding and exploiting language diversity.
\newblock In {\em International Joint Conference on Artificial Intelligence (IJCAI)}, pages 4009--4017.

\bibitem[Giunchiglia et~al., 2023]{giunchiglia2023representing}
Giunchiglia, F., Bella, G., Nair, N.~C., Chi, Y., and Xu, H. (2023).
\newblock Representing interlingual meaning in lexical databases.
\newblock {\em Artificial Intelligence Review}, 56(10):11053--11069.

\bibitem[Giunchiglia et~al., 2015]{giunchiglia2015}
Giunchiglia, F., Jovanovic, M., Huertas-Miguel{\'a}{\~n}ez, M., Batsuren, K., et~al. (2015).
\newblock Crowdsourcing a large scale multilingual lexico-semantic resource.
\newblock In {\em AAAI Conference on Human Computation and Crowdsourcing (HCOMP-15)}.

\bibitem[Hasanain et~al., 2024]{hasanain2023}
Hasanain, M., Ahmad, F., and Alam, F. (2024).
\newblock Large language models for propaganda span annotation.

\bibitem[Kasumba and Neumman, 2024]{kasumba2024}
Kasumba, R. and Neumman, M. (2024).
\newblock Practical sentiment analysis for education: The power of student crowdsourcing.
\newblock {\em Proceedings of the AAAI Conference on Artificial Intelligence}, 38(21):23110--23118.

\bibitem[Katsuta and Yamamoto, 2020]{katsuta2020l}
Katsuta, A. and Yamamoto, K. (2020).
\newblock Lexical simplification by unsupervised machine translation.
\newblock {\em International Journal of Asian Language Processing}, 30(02):2050008.

\bibitem[Kay and Cook, 2016]{kay2016}
Kay, P. and Cook, R.~S. (2016).
\newblock World color survey.
\newblock In Luo, M.~R., editor, {\em Encyclopedia of Color Science and Technology}, pages 1265--1271. Springer, New York.

\bibitem[Kemp and Regier, 2012]{kemp2012}
Kemp, C. and Regier, T. (2012).
\newblock Kinship categories across languages reflect general communicative principles.
\newblock {\em Science}, 336(6084):1049--1054.

\bibitem[Khalilia et~al., 2023]{Khalilia2023}
Khalilia, H., Bella, G., Freihat, A.~A., Darma, S., and Giunchiglia, F. (2023).
\newblock Lexical diversity in kinship across languages and dialects.
\newblock {\em Frontiers in Psychology}, 14.

\bibitem[Khishigsuren et~al., 2022]{khishigsuren2022}
Khishigsuren, T., Bella, G., Batsuren, K., Freihat, A.~A., Chandran~Nair, N., Ganbold, A., Khalilia, H., Chandrashekar, Y., and Giunchiglia, F. (2022).
\newblock Using linguistic typology to enrich multilingual lexicons: the case of lexical gaps in kinship.
\newblock In Calzolari, N., B{\'e}chet, F., Blache, P., Choukri, K., Cieri, C., Declerck, T., Goggi, S., Isahara, H., Maegaard, B., Mariani, J., Mazo, H., Odijk, J., and Piperidis, S., editors, {\em Proceedings of the Thirteenth Language Resources and Evaluation Conference}, pages 2798--2807, Marseille, France. European Language Resources Association.

\bibitem[Kirby et~al., 2016]{kirby2016}
Kirby, K.~R., Gray, R.~D., Greenhill, S.~J., Jordan, F.~M., Gomes-Ng, S., Bibiko, H.-J., Blasi, D.~E., Botero, C.~A., Bowern, C., Ember, C.~R., et~al. (2016).
\newblock {D-PLACE}: A global database of cultural, linguistic and environmental diversity.
\newblock {\em PLOS ONE}, 11(7):e0158391.

\bibitem[Kopecka and Narasimhan, 2012]{kopecka2012}
Kopecka, A. and Narasimhan, B. (2012).
\newblock {\em Events of putting and taking: A crosslinguistic perspective}, volume 100.
\newblock John Benjamins Publishing.

\bibitem[Koto et~al., 2020]{koto-etal-2020-indolem}
Koto, F., Rahimi, A., Lau, J.~H., and Baldwin, T. (2020).
\newblock {I}ndo{LEM} and {I}ndo{BERT}: A benchmark dataset and pre-trained language model for {I}ndonesian {NLP}.
\newblock In Scott, D., Bel, N., and Zong, C., editors, {\em Proceedings of the 28th International Conference on Computational Linguistics}, pages 757--770, Barcelona, Spain (Online). International Committee on Computational Linguistics.

\bibitem[Krippendorff, 2011]{krippendorff2011}
Krippendorff, K. (2011).
\newblock Computing krippendorff’s alpha-reliability.

\bibitem[Lang, 2001]{Lang2001}
Lang, E. (2001).
\newblock Spatial dimension terms.
\newblock In {\em Language Typology and Language Universals}, volume~2, pages 1251--1275. De Gruyter Mouton, Berlin, Boston.

\bibitem[Lanser et~al., 2016]{lanser2016}
Lanser, B., Unger, C., and Cimiano, P. (2016).
\newblock Crowdsourcing ontology lexicons.
\newblock In Calzolari, N., Choukri, K., Declerck, T., Goggi, S., Grobelnik, M., Maegaard, B., Mariani, J., Mazo, H., Moreno, A., Odijk, J., and Piperidis, S., editors, {\em Proceedings of the Tenth International Conference on Language Resources and Evaluation ({LREC}'16)}, pages 3477--3484, Portoro{\v{z}}, Slovenia. European Language Resources Association (ELRA).

\bibitem[Lease and Yilmaz, 2012]{lease2012}
Lease, M. and Yilmaz, E. (2012).
\newblock Crowdsourcing for information retrieval.
\newblock {\em SIGIR Forum}, 45(2):66–75.

\bibitem[Lehrer, 1970]{lehrer1970}
Lehrer, A. (1970).
\newblock Notes on lexical gaps.
\newblock {\em Journal of Linguistics}, 6(2):257--261.

\bibitem[Levinson, 2000]{levinson2000presumptive}
Levinson, S.~C. (2000).
\newblock {\em Presumptive meanings: The theory of generalized conversational implicature}.
\newblock MIT press.

\bibitem[Lin et~al., 2024]{lin2024}
Lin, Z., Chen, W., Song, Y., and Zhang, Y. (2024).
\newblock Prompting few-shot multi-hop question generation via comprehending type-aware semantics.
\newblock In {\em Findings of the Association for Computational Linguistics: NAACL 2024}, pages 3730--3740.

\bibitem[Liu et~al., 2013]{liu2013}
Liu, Q., Ihler, A.~T., and Steyvers, M. (2013).
\newblock Scoring workers in crowdsourcing: How many control questions are enough?
\newblock In Burges, C., Bottou, L., Welling, M., Ghahramani, Z., and Weinberger, K., editors, {\em Advances in Neural Information Processing Systems}, volume~26. Curran Associates, Inc.

\bibitem[Loureiro and Jorge, 2019]{loureiro2019}
Loureiro, D. and Jorge, A. (2019).
\newblock Language modelling makes sense: Propagating representations through wordnet for full-coverage word sense disambiguation.
\newblock {\em arXiv preprint arXiv:1906.10007}.

\bibitem[Majid et~al., 2007]{majid2007}
Majid, A., Bowerman, M., van Staden, M., and Boster, J.~S. (2007).
\newblock The semantic categories of cutting and breaking events: A crosslinguistic perspective.
\newblock {\em Cognitive Linguistics}, 18(2):133--152.

\bibitem[Manerkar et~al., 2022]{manerkar2022}
Manerkar, S., Asnani, K., Khorjuvenkar, P.~R., Desai, S., and Pawar, J.~D. (2022).
\newblock Konkani wordnet: Corpus-based enhancement using crowdsourcing.
\newblock {\em Transactions on Asian and Low-Resource Language information Processing}, 21(4):1--18.

\bibitem[McCarthy et~al., 2019]{mccarthy2019}
McCarthy, A.~D., Wu, W., Mueller, A., Watson, B., and Yarowsky, D. (2019).
\newblock Modeling color terminology across thousands of languages.
\newblock {\em arXiv preprint arXiv:1910.01531}.

\bibitem[Mikolov et~al., 2013]{mikolov2013efficientestimationwordrepresentations}
Mikolov, T., Chen, K., Corrado, G., and Dean, J. (2013).
\newblock Efficient estimation of word representations in vector space.

\bibitem[Miller, 1995]{miller1995}
Miller, G.~A. (1995).
\newblock Wordnet: a lexical database for english.
\newblock {\em Communications of the ACM}, 38(11):39--41.

\bibitem[Murdock, 1970]{murdock1970}
Murdock, G.~P. (1970).
\newblock Kin term patterns and their distribution.
\newblock {\em Ethnology}, 9(2):165--208.

\bibitem[Nair, 2022]{nair2022}
Nair, N. (2022).
\newblock A crowdsourcing methodology for improving the malayalam wordnet.
\newblock {\em Available at SSRN 4064783}.

\bibitem[Noor et~al., 2011]{noor2011}
Noor, N. H. B.~M., Sapuan, S., and Bond, F. (2011).
\newblock Creating the open {W}ordnet {B}ahasa.
\newblock In {\em Proceedings of the 25th Pacific Asia Conference on Language, Information and Computation}, pages 255--264. Institute of Digital Enhancement of Cognitive Processing, Waseda University.

\bibitem[Ono et~al., 2023]{ono2023consideration}
Ono, M., Soga, T., Kikuchi, M., and Tanabe, T. (2023).
\newblock Consideration of language learning service with visualized vocabulary map derived from wordnet.
\newblock In {\em 2023 8th International Conference on Business and Industrial Research (ICBIR)}, pages 1194--1198. IEEE.

\bibitem[Parent and Eskenazi, 2010]{parent2010}
Parent, G. and Eskenazi, M. (2010).
\newblock Clustering dictionary definitions using {A}mazon {M}echanical {T}urk.
\newblock In Callison-Burch, C. and Dredze, M., editors, {\em Proceedings of the {NAACL} {HLT} 2010 Workshop on Creating Speech and Language Data with {A}mazon{'}s Mechanical Turk}, pages 21--29, Los Angeles. Association for Computational Linguistics.

\bibitem[Pianta et~al., 2002]{pianta2002}
Pianta, E., Bentivogli, L., and Girardi, C. (2002).
\newblock Developing an aligned multilingual database.
\newblock In {\em Proceedings of the 1st International WordNet Conference}, pages 293--302. Global Wordnet Association.

\bibitem[Plungyan, 2011]{plungyan2011}
Plungyan, V. (2011).
\newblock Modern linguistic typology.
\newblock {\em Herald of the Russian Academy of Sciences}, 81(2):101--113.

\bibitem[Post et~al., 2012]{post2012}
Post, M., Callison-Burch, C., and Osborne, M. (2012).
\newblock Constructing parallel corpora for six {I}ndian languages via crowdsourcing.
\newblock In Callison-Burch, C., Koehn, P., Monz, C., Post, M., Soricut, R., and Specia, L., editors, {\em Proceedings of the Seventh Workshop on Statistical Machine Translation}, pages 401--409, Montr{\'e}al, Canada. Association for Computational Linguistics.

\bibitem[Powers, 2012]{powers2012}
Powers, D. M.~W. (2012).
\newblock The problem with kappa.
\newblock In Daelemans, W., editor, {\em Proceedings of the 13th Conference of the {E}uropean Chapter of the Association for Computational Linguistics}, pages 345--355, Avignon, France. Association for Computational Linguistics.

\bibitem[Reznikova et~al., 2012]{reznikova2012}
Reznikova, T., Rakhilina, E., and Bonch-Osmolovskaya, A. (2012).
\newblock Towards a typology of pain predicates.
\newblock {\em Linguistics}, 50(3):421--465.

\bibitem[Roberson et~al., 2005]{roberson2005}
Roberson, D., Davidoff, J., Davies, I.~R., and Shapiro, L.~R. (2005).
\newblock Color categories: Evidence for the cultural relativity hypothesis.
\newblock {\em Cognitive Psychology}, 50(4):378--411.

\bibitem[Robinson et~al., 2019]{robinson2019}
Robinson, J., Rosenzweig, C., Moss, A.~J., and Litman, L. (2019).
\newblock Tapped out or barely tapped? recommendations for how to harness the vast and largely unused potential of the mechanical turk participant pool.
\newblock {\em PloS one}, 14(12):e0226394.

\bibitem[Sayeed et~al., 2024]{sayeed2024annotating}
Sayeed, H.~M., Mohanty, T., and Sparks, T.~D. (2024).
\newblock Annotating materials science text: A semi-automated approach for crafting outputs with gemini pro.
\newblock {\em Integrating Materials and Manufacturing Innovation}, 13(2):445--452.

\bibitem[Sneddon, 2003]{sneddon2003}
Sneddon, J. (2003).
\newblock {\em The Indonesian Language}.
\newblock University of New South Wales Press Ltd, Sydney.

\bibitem[Team et~al., 2022]{nllbteam2022languageleftbehindscaling}
Team, N., Costa-jussà, M.~R., Cross, J., Çelebi, O., Elbayad, M., Heafield, K., Heffernan, K., Kalbassi, E., Lam, J., Licht, D., Maillard, J., Sun, A., Wang, S., Wenzek, G., Youngblood, A., Akula, B., Barrault, L., Gonzalez, G.~M., Hansanti, P., Hoffman, J., Jarrett, S., Sadagopan, K.~R., Rowe, D., Spruit, S., Tran, C., Andrews, P., Ayan, N.~F., Bhosale, S., Edunov, S., Fan, A., Gao, C., Goswami, V., Guzmán, F., Koehn, P., Mourachko, A., Ropers, C., Saleem, S., Schwenk, H., and Wang, J. (2022).
\newblock No language left behind: Scaling human-centered machine translation.

\bibitem[{Tim Penyusun Kamus Pusat Bahasa}, 2008]{kbbi}
{Tim Penyusun Kamus Pusat Bahasa} (2008).
\newblock {\em Kamus Bahasa Indonesia}.
\newblock Pusat Bahasa Departemen Pendidikan Nasional.

\bibitem[T{\"o}rnberg, 2023]{tornberg2023}
T{\"o}rnberg, P. (2023).
\newblock Chatgpt-4 outperforms experts and crowd workers in annotating political twitter messages with zero-shot learning.

\bibitem[W{\"a}lchli and Cysouw, 2012]{walchli2012}
W{\"a}lchli, B. and Cysouw, M. (2012).
\newblock Lexical typology through similarity semantics: Toward a semantic map of motion verbs.
\newblock {\em Linguistics}, 50(3):671--710.

\bibitem[Warrens, 2011]{warrens2011}
Warrens, M.~J. (2011).
\newblock Cohen’s kappa is a weighted average.
\newblock {\em Statistical Methodology}, 8(6):473--484.

\bibitem[Welbl et~al., 2017]{welbl2017}
Welbl, J., Liu, N.~F., and Gardner, M. (2017).
\newblock Crowdsourcing multiple choice science questions.
\newblock {\em arXiv preprint arXiv:1707.06209}.

\bibitem[Wierzbicka, 1996]{wierzbicka1996semantics}
Wierzbicka, A. (1996).
\newblock {\em Semantics: Primes and universals: Primes and universals}.
\newblock Oxford University Press, UK.

\bibitem[Wierzbicka, 2007]{wierzbicka2007}
Wierzbicka, A. (2007).
\newblock Bodies and their parts: An nsm approach to semantic typology.
\newblock {\em Language Sciences}, 29(1):14--65.

\bibitem[Wijesiri et~al., 2014]{wijesiri2014}
Wijesiri, I., Gallage, M., Gunathilaka, B., Lakjeewa, M., Wimalasuriya, D., Dias, G., Paranavithana, R., and de~Silva, N. (2014).
\newblock Building a {W}ord{N}et for {S}inhala.
\newblock In Orav, H., Fellbaum, C., and Vossen, P., editors, {\em Proceedings of the Seventh Global {W}ordnet Conference}, pages 100--108, Tartu, Estonia. University of Tartu Press.

\bibitem[Åke Viberg, 1984]{viberg1983}
Åke Viberg (1984).
\newblock {\em The verbs of perception: a typological study}, pages 123--162.
\newblock De Gruyter Mouton, Berlin, Boston.

\end{thebibliography}






\end{document}